\documentclass{article} 
\usepackage{iclr2021_conference,times}

\usepackage{etex}  
\usepackage{standalone}
\usepackage{latexsym}
\usepackage{url}
\usepackage{amsmath}
\usepackage{amssymb}
\usepackage{amsfonts}
\usepackage{amsthm}
\usepackage{graphicx}
\usepackage{subcaption}
\usepackage{array}
\usepackage{tabu}
\usepackage{makecell}
\usepackage{subcaption}
\usepackage{paralist}
\usepackage{cases}
\usepackage{diagbox}
\usepackage{enumitem}
\usepackage{soul}
\usepackage{multirow}
\usepackage{verbatim}
\usepackage{tabulary}
\usepackage{booktabs}
\usepackage[mathscr]{euscript}
\usepackage{mathtools}
\usepackage{algorithm}
\usepackage{algpseudocode}
\usepackage{stmaryrd}
\usepackage{tikz-dependency}
\usepackage{subcaption}
\usetikzlibrary{automata,decorations.markings,arrows,positioning,matrix,calc,patterns,angles,quotes,calc}
\usepackage{adjustbox}
\usepackage{tabularx}
\usepackage{xspace}
\usepackage{tabulary}
\usepackage{afterpage}
\usepackage{hyperref}
\usepackage{url}
\usepackage{bm}
\usepackage{color}
\usepackage{graphicx}
\usepackage{slashbox}
\usepackage[toc,page]{appendix}

\usepackage{amsmath,amsfonts,bm}

\def\eqref#1{equation~\ref{#1}}

\def\1{\bm{1}}

\def\vzero{{\mathbf{0}}}

\def\vh{{\mathbf{h}}}

\def\vk{{\mathbf{k}}}

\def\vq{{\mathbf{q}}}

\def\vv{{\mathbf{v}}}
\def\vw{{\mathbf{w}}}
\def\vx{{\mathbf{x}}}
\def\vy{{\mathbf{y}}}
\def\vz{{\mathbf{z}}}

\def\mI{{\mathbf{I}}}

\def\mS{{\mathbf{S}}}

\DeclareMathAlphabet{\mathsfit}{\encodingdefault}{\sfdefault}{m}{sl}
\SetMathAlphabet{\mathsfit}{bold}{\encodingdefault}{\sfdefault}{bx}{n}

\def\gN{{\mathcal{N}}}
\def\gO{{\mathcal{O}}}

\definecolor{orange}{rgb}{1,0.5,0}
\definecolor{mdgreen}{rgb}{0.05,0.6,0.05}
\definecolor{mdblue}{rgb}{0,0,0.7}
\definecolor{dkblue}{rgb}{0,0,0.5}
\definecolor{dkgray}{rgb}{0.3,0.3,0.3}
\definecolor{slate}{rgb}{0.25,0.25,0.4}
\definecolor{gray}{rgb}{0.5,0.5,0.5}
\definecolor{ltgray}{rgb}{0.7,0.7,0.7}
\definecolor{purple}{rgb}{0.7,0,1.0}
\definecolor{lavender}{rgb}{0.65,0.55,1.0}

\definecolor{mypurple}{RGB}{111,61,121}
\definecolor{myblue}{RGB}{46,88,180}
\definecolor{myred}{RGB}{181,68,106}
\definecolor{myyellow}{RGB}{204,143,55}

\newcommand{\ensuretext}[1]{#1}

\newcommand{\arkcomment}[3]{\ensuretext{\textcolor{#3}{[#1 #2]}}}
\renewcommand{\arkcomment}[3]{}  

\newcommand{\rev}[1]{{#1}}

\newcommand{\term}[1]{\textbf{#1}} 

\newcommand{\interalia}[1]{\citep[\emph{inter alia}]{#1}}

\newcommand{\relu}{\operatorname{ReLU}}

\newcommand{\norm}[1]{\left\lVert#1\right\rVert}
\DeclareSymbolFont{extraup}{U}{zavm}{m}{n}
\DeclareMathSymbol{\vardiamond}{\mathalpha}{extraup}{87}

\newcolumntype{L}[1]{>{\raggedright\let\newline\\\arraybackslash\hspace{0pt}}m{#1}}
\newcolumntype{C}[1]{>{\centering\let\newline\\\arraybackslash\hspace{0pt}}m{#1}}
\newcolumntype{R}[1]{>{\raggedleft\let\newline\\\arraybackslash\hspace{0pt}}m{#1}}

\newtheorem{theorem}{Theorem}

\theoremstyle{definition}

\theoremstyle{remark}

\algrenewcommand{\algorithmiccomment}[1]{\leavevmode$\triangleright$ #1}

\setul{1pt}{.4pt}

\DeclareCaptionFont{tiny}{\tiny}

\usepackage{url}
\usepackage[shortcuts]{extdash}  

\usepackage{blindtext}
\usepackage{graphicx}
\usepackage{capt-of}
\usepackage{booktabs}
\usepackage{varwidth}
\newsavebox\tmpbox

\title{Random Feature Attention}

\author{Hao Peng$^\spadesuit$\thanks{The majority of this
work was done while these authors were at DeepMind.} \quad
        Nikolaos Pappas$^\spadesuit$ \quad
        Dani Yogatama$^\clubsuit$ \quad
        Roy Schwartz$^\heartsuit$ \\
        \textbf{Noah A. Smith}$^{\spadesuit\diamondsuit}$\quad
        \textbf{Lingpeng Kong}$^{\vardiamond\ast}$\\
  $^\spadesuit$Paul G. Allen School of Computer Science \& Engineering,
  University of Washington\\
  $^\clubsuit$DeepMind\quad
  $^\diamondsuit$Allen Institute for Artificial Intelligence\\ 
  $^\heartsuit$School of Computer Science \& Engineering, Hebrew University of Jerusalem\\
  $^\vardiamond$Department of Computer Science , The University of Hong Kong \\ 
  {\tt \{hapeng,npappas,nasmith\}@cs.washington.edu} \\
  {\tt dyogatama@google.com,
  roys@cs.huji.ac.il,
  lpk@cs.hku.hk}
}

\newcommand{\model}{\textsc{Rfa}\xspace}
\newcommand{\modelgate}{\textsc{Rfa-Gate}\xspace}

\newcommand{\resolved}[1]{}
\newcommand{\base}[0]{\textsc{Base}\xspace}
\iclrfinalcopy 
\begin{document}

\maketitle
\begin{abstract}
    Transformers are state-of-the-art models
    for a variety of sequence modeling tasks.
    At their core is an attention function 
    which models pairwise interactions
    between the inputs at every timestep. 
    While attention is powerful, 
    it does \emph{not} scale efficiently to long sequences
    due to its quadratic time and space complexity in the sequence length.
    We propose \model, 
    a linear time and space
    \textbf{a}ttention that uses \textbf{r}andom \textbf{f}eature methods
    to approximate the softmax function, and explore its application in transformers.
    \model can be used as a drop-in replacement for conventional softmax attention and offers a straightforward way of learning with recency bias
    through an optional gating mechanism.
    Experiments on language modeling and machine translation
    demonstrate that \model
    achieves similar or better
    performance compared to strong transformer baselines.
    In the machine translation experiment,
    \model decodes twice as fast as a vanilla transformer.
    Compared to existing efficient transformer variants, 
    \model is competitive in terms of both accuracy and efficiency on three long text classification datasets.
    Our analysis shows that 
    \model's efficiency gains are especially notable on long sequences,
    suggesting that \model will be particularly useful
    in tasks that require
    working with large inputs, fast decoding speed, or low memory footprints. 

\end{abstract}
\section{Introduction}
Transformer architectures~\citep{vaswani2017attention}
have achieved tremendous success on a variety of sequence modeling
tasks~\interalia{ott2018scaling,radford2018language,parmar2018image,delvin2019bert,parisotto2019stabilizing}.
Under the hood, the key component is attention \citep{bahdanau2015attention}, which models pairwise interactions of the inputs,
regardless of their distances from each other.
This comes with quadratic time and memory costs,
making the transformers computationally expensive, especially for long sequences.
A large body of research has been devoted to
improving their time and memory efficiency~\citep{tay2020}.
Although better \emph{asymptotic} complexity
and prominent gains for long sequences have been achieved~\interalia{lee2019set,child2019generating,beltagy2020longformer},
in practice, many existing approaches are less well-suited
for moderate-length ones:
the additional computation steps required by some approaches can  overshadow
the time and memory they save~\interalia{kitaev2020reformer,wang2020linformer,roy2020efficient}.

This work proposes \textbf{r}andom \textbf{f}eature
\textbf{a}ttention (\model),
\rev{an efficient attention variant}
that scales linearly in sequence length
in terms of time and space,
\rev{and achieves practical gains for both long and moderate length sequences.}
\model builds on a kernel perspective of softmax~\citep{rawat2019sampled}.
Using the well-established random feature maps (\citealp{rahimi2009rff,avron2016qmc}; \S\ref{sec:background}),
\model approximates the dot-then-exponentiate function 
with a kernel trick~\citep{hofmann2008kernel}: ${\exp(\vx\cdot\vy)
\approx\boldsymbol{\phi}(\vx)\cdot\boldsymbol{\phi}(\vy)}$.
Inspired by its connections to
gated recurrent neural networks~\citep{hochreiter1997lstm,cho2014gru}
and  fast weights \citep{schmidhuber1992learning},
we further augment \model with an optional gating mechanism,
offering a straightforward way of learning with recency bias
when locality is desired.

\model and its gated variant (\S\ref{sec:model})
can be used as a drop-in substitute for the canonical softmax attention,
and increase the number of parameters by less than 0.1\%.
We explore its applications in transformers
on language modeling, machine translation, \rev{and long text classification} (\S\ref{sec:experiments}).
Our experiments
show that \model achieves comparable performance to vanilla transformer baselines
in \rev{all} tasks, while outperforming a recent related approach \citep{katharopoulos20transformers}.
The gating mechanism proves
particularly useful in language modeling: the gated variant of \model
outperforms the transformer baseline on WikiText-103.
\model shines in decoding, even for shorter sequences.
In our head-to-head comparison on machine translation benchmarks, 
\model decodes around $2\times$ faster than a transformer baseline,
\emph{without} accuracy loss. 
\rev{Comparisons to several recent efficient transformer variants on three 
long text classification datasets show that
\model is competitive in terms of both accuracy and efficiency.
}
Our analysis (\S\ref{sec:analysis}) shows that more significant time and memory efficiency
improvements
can be achieved for longer sequences: 12$\times$ decoding speedup with less than 10\% of the memory for 2,048-length outputs.

\section{Background}\label{sec:background}

\subsection{Attention in Sequence Modeling}\label{sec:attention}

The attention mechanism~\citep{bahdanau2015attention} has been widely used in many sequence modeling tasks.
Its dot-product variant is the key building block for the 
state-of-the-art transformer architectures~\citep{vaswani2017attention}.
Let $\{\vq_t\}_{t=1}^{N}$ denote
a sequence of $N$ \term{query} vectors,
that attend to sequences of $M$ \term{key} and \term{value} vectors.
At each timestep, the attention
linearly combines the values weighted by the outputs of a softmax:
\begin{align}\label{eq:attention}
\begin{split}
    \operatorname{attn}\left(\vq_t,\{\vk_i\},\{\vv_i\}\right)=\sum_i
    \frac{\exp\left(\vq_t\cdot\vk_i /\tau\right)}
    {\sum_j\exp\left(\vq_t\cdot\vk_j /\tau\right)}\vv_i^\top.
\end{split}
\end{align}
$\tau$ is the temperature hyperparameter determining how ``flat'' the softmax is~\citep{hinton2015distilling}.\footnote{
 $M=N$ in self-attention; they may differ, e.g., in the cross attention of a sequence-to-sequence model.
}

Calculating attention 
for a single query takes $\gO(M)$ time and space.
For the full sequence of $N$ queries
the space amounts to $\gO(MN)$.
When the computation \emph{cannot} be parallelized 
across the queries, e.g., in autoregressive decoding,
the time complexity is quadratic in the sequence length.

\subsection{Random Feature Methods}\label{sec:random_feature}
The theoretical backbone of this work is the 
unbiased estimation of the Gaussian kernel by \citet{rahimi2009rff}.
Based on Bochner's theorem~\citep{bochner1955harmonic},
\citet{rahimi2009rff} proposed random Fourier features to approximate  a desired shift-invariant kernel. The method nonlinearly transforms a pair of vectors $\vx$ and $\vy$ using a \term{random feature map} $\boldsymbol{\phi}$;
the inner product between $\boldsymbol{\phi}(\vx)$ and $\boldsymbol{\phi}(\vy)$
approximates the kernel evaluation on $\vx$ and $\vy$.
More precisely:
\begin{theorem}[\citealp{rahimi2009rff}]\label{thm:bochner}
Let $\boldsymbol{\phi}: \mathbb{R}^d \rightarrow\mathbb{R}^{2D}$ be a nonlinear transformation:
\begin{align}\label{eq:bochner}
    \boldsymbol{\phi}\left(\vx\right) = \sqrt{1/D}
    \Bigl[
        \sin\left(\vw_1\cdot\vx\right),
        \dots,
        \sin\left(\vw_D\cdot\vx\right),
        \cos\left(\vw_1\cdot\vx\right),
        \dots,
        \cos\left(\vw_D\cdot\vx\right)
    \Bigr]^\top.
\end{align}
When $d$-dimensional random vectors $\vw_i$
are independently sampled from $\gN(\vzero, \sigma^2\mI_d)$,
\begin{align}\label{eq:approx_gaussian_kernel}
\mathbb{E}_{\vw_i}\left[\boldsymbol{\phi}\left(\vx\right)\cdot \boldsymbol{\phi}\left(\vy\right)\right]
= \exp\left({-\norm{\vx-\vy}^2/2\sigma^2}\right).
\end{align}
\end{theorem}%
Variance of the estimation 
is inversely proportional to $D$~(\rev{Appendix~\ref{sec:variance}};~\citealp{yu2016orf}).

Random feature methods proved successful in speeding up kernel methods~\interalia{junier2015fast,avron2017faster,sun2019random},
and more recently are used to efficiently approximate softmax~\citep{rawat2019sampled}.
In \S\ref{sec:rfa}, we use it 
to derive an unbiased estimate to $\exp(\langle\boldsymbol{\cdot}\,,\boldsymbol{\cdot}\rangle)$
and then an efficient approximation to softmax attention.

\section{Model}
\label{sec:model} 

This section presents \model (\S\ref{sec:rfa}) 
and its gated variant (\S\ref{sec:gate}).
In \S\ref{sec:discussion}
we lay out several design choices
and relate \model to prior works.
We close by practically analyzing
\model's complexity~(\S\ref{sec:complexity}).

\subsection{Random Feature Attention}\label{sec:rfa}

\model builds on an unbiased estimate to $\exp(\langle\boldsymbol{\cdot}\,,\boldsymbol{\cdot}\rangle)$
from Theorem~\ref{thm:bochner}, which we begin with:
\begin{align}\label{eq:approx_dot_exp}
\begin{split}
    \exp\left({\vx\cdot\vy / \sigma^2} \right)
    &= \exp \left({\norm{\vx}^2/2\sigma^2 + \norm{\vy}^2/2\sigma^2}\right) \exp\left({-\norm{\vx-\vy}^2 / 2\sigma^2} \right)\\
    &\approx \exp \left({\norm{\vx}^2/2\sigma^2 + \norm{\vy}^2/2\sigma^2}\right)\ 
    \boldsymbol{\phi}\left(\vx\right)\cdot \boldsymbol{\phi}\left(\vy\right).
\end{split}
\end{align}
The last line
does \emph{not} have any nonlinear interaction between $\boldsymbol{\phi}(\vx)$ and $\boldsymbol{\phi}(\vy)$,
allowing for a linear time/space approximation 
to attention.
For clarity we assume the query and keys are unit vectors.\footnote{
This can be achieved by $\ell_2$-normalizing the query and keys.
See \S\ref{sec:discussion} for a related discussion.} 
\begin{align}\label{eq:approx_softmax}
\begin{split}
    \operatorname{attn}\left(\vq_t,\{\vk_i\},\{\vv_i\}\right)
    &=\sum_i
    \frac{\exp\left(\vq_t\cdot\vk_i /\sigma^2\right)}
    {\sum_j\exp\left(\vq_t\cdot\vk_j /\sigma^2\right)}\vv_i^\top\\
    &\approx 
    \sum_i\frac{\boldsymbol{\phi}\left(\vq_t\right)^\top \boldsymbol{\phi}\left(\vk_i\right)\vv_i^\top}
    {\sum_j\boldsymbol{\phi}\left(\vq_t\right) \cdot \boldsymbol{\phi}\left(\vk_j\right)}\\
    &= \frac{\boldsymbol{\phi}\left(\vq_t\right)^\top\sum_i \boldsymbol{\phi}\left(\vk_i\right)\otimes\vv_i}
    {\boldsymbol{\phi}\left(\vq_t\right)\cdot\sum_j \boldsymbol{\phi}\left(\vk_j\right)}
    =\model\left(\vq_t,\{\vk_i\},\{\vv_i\}\right).
\end{split}
\end{align} 
$\otimes$ denotes the outer product between vectors, 
and $\sigma^2$ corresponds to the temperature term $\tau$ in Eq.~\ref{eq:attention}.

\begin{figure}[t]
\centering
\begin{subfigure}[tb]{0.48\textwidth}
\includegraphics[trim={.0cm 0cm 0cm 0cm},clip,width=\textwidth]{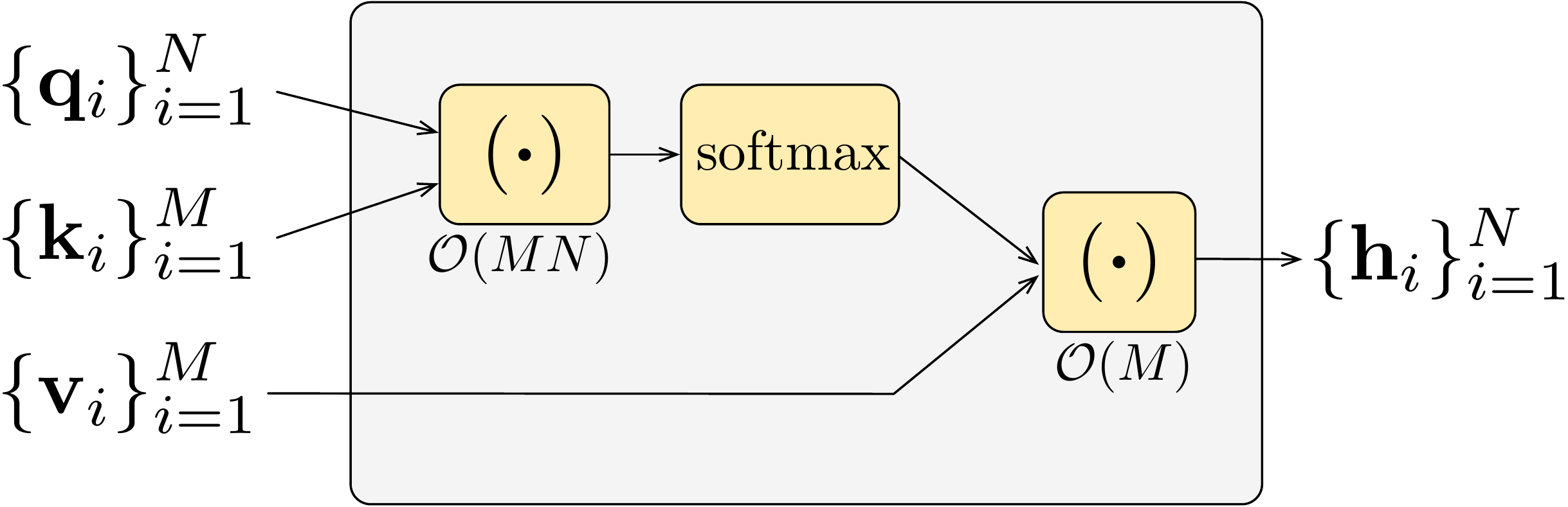}
\caption{\label{fig:softmax} Softmax attention.}
\end{subfigure}
\hfill
\begin{subfigure}[tb]{0.48\textwidth}
\includegraphics[trim={.0cm 0cm 0cm 0cm},clip,width=\textwidth]{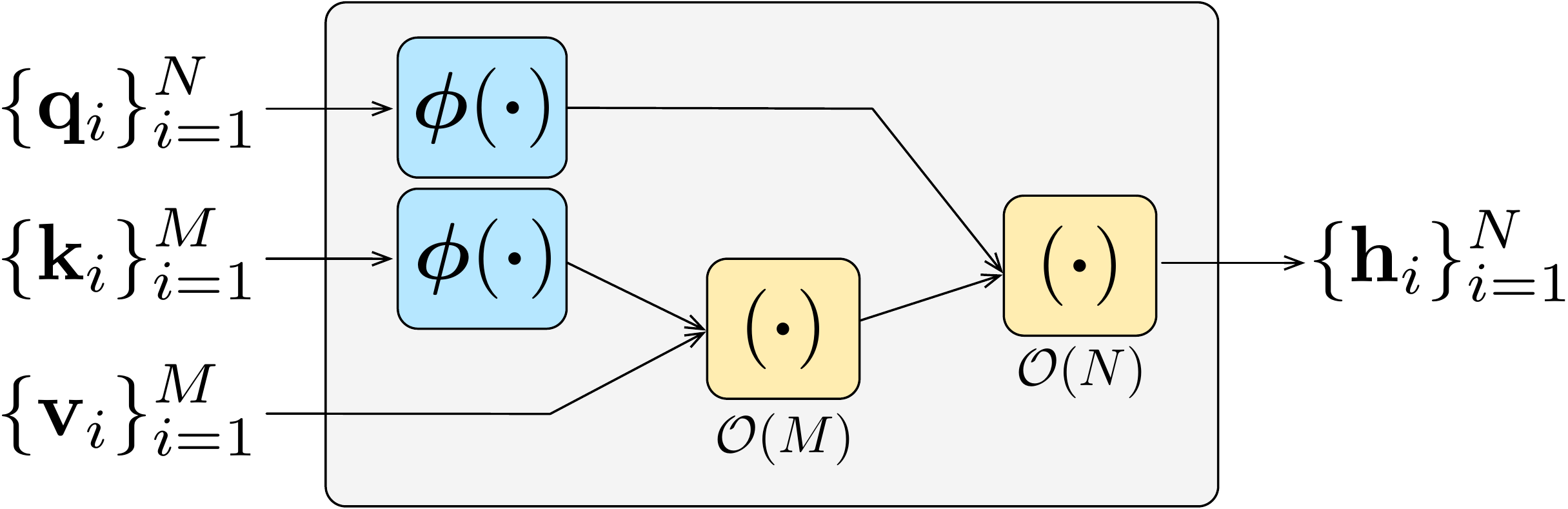}
\caption{\label{fig:rfa}Random feature attention.}
\end{subfigure}
\caption{Computation graphs for softmax attention (left) and random feature attention (right).
Here, we assume cross attention with source length $M$
and target length $N$.
}\label{fig:comp_graph}
\end{figure}

\model can be used as a drop-in-replacement for softmax-attention. 
\begin{compactitem}
\item[(a)] The input is revealed in full to \term{cross attention}
and \term{encoder self-attention}. 
\rev{Here \model calculates attention using Eq.~\ref{eq:approx_softmax}}.
\item[(b)]
In \term{causal attention} \model attends only to the prefix.\footnote{
It is also sometimes called ``decoder self-attention'' or ``autoregressive attention.''
}
This allows for a recurrent computation.
Tuple $(\mS_t\in\mathbb{R}^{2D\times d},\vz_t\in\mathbb{R}^{2D})$
is used as the ``hidden state'' at time step $t$ to keep track of the history, similar to those in RNNs.
Then
$\model(\vq_t,\{\vk_i\}_{i\leq t},\{\vv_i\}_{i\leq t}) = \boldsymbol{\phi}(\vq_t)^\top\mS_t / (\boldsymbol{\phi}(\vq_t)\cdot\vz_t)$, 
where
\begin{align}\label{eq:linear_causal}
    \mS_t = \mS_{t-1} +  \boldsymbol{\phi}\left(\vk_t\right)\otimes\vv_t, \quad
    \vz_t = \vz_{t-1} + \boldsymbol{\phi}\left(\vk_t\right).
\end{align}
\end{compactitem}
\rev{$2D$ denotes the size of $\boldsymbol{\phi}(\boldsymbol{\cdot}).$}
Appendix~\ref{sec:algo} summarizes the computation procedure of \model,
and Figure~\ref{fig:comp_graph} compares it
against the softmax attention.
Appendix~\ref{sec:causal_rfa_derivation} derives causal \model in detail.

Analogously to the softmax attention, \model has its multiheaded variant~\citep{vaswani2017attention}. 
In our experiments we use causal \model in a transformer language model (\S\ref{sec:lm}), and both cross and causal \model in the decoder of a sequence-to-sequence machine translation model.

\subsection{\model-Gate: Learning with Recency Bias}\label{sec:gate}
The canonical softmax attention does \emph{not} have 
any explicit modeling of distance or locality.
In learning problems where such inductive bias
is crucial~\interalia{
ba2016using,parmar2018image,thomas2018differentiable,li2019enhancing},
transformers heavily rely on positional encodings.
Answering to this, many approaches have been proposed,
e.g.,
learning the attention spans~\citep{sukhbaatar2019adaptive,wu2020Lite},
and enhancing the attention computation with recurrent~\citep{hao2019modeling,chen2019recurrent}
or convolutional~\citep{wu2018pay,mohamed2019transformers} components.

\model faces the same issue, but its causal attention variant (Eq.~\ref{eq:linear_causal})
offers a straightforward way of learning with recency bias.
We draw inspiration from its connections to RNNs, 
and augment \model with a learned gating mechanism~\interalia{hochreiter1997lstm,cho2014gru,peng2018rational}:
\begin{align}\label{eq:gate_rfa}
\begin{split}
    g_t &= \operatorname{sigmoid}( \vw_g\cdot \vx_t+ b_g), \\
    \mS_t &= g_t\, \mS_{t-1} +  (1-g_t)\,\boldsymbol{\phi}\left(\vk_t\right)\otimes\vv_t, \\
    \vz_t &= g_t\, \vz_{t-1} + (1-g_t)\,\boldsymbol{\phi}\left(\vk_t\right).
\end{split}
\end{align}
$\vw_g$ and $b_g$ are learned parameters,
and $\vx_t$ is the input representation at timestep $t$.\footnote{
In multihead attention~\citep{vaswani2017attention},
$\vk_t$ and $\vv_t$ are calculated from $\vx_t$
using learned affine transformations.
}
By multiplying the learned scalar gates $0< g_t<1$ against the 
hidden state $(\mS_t, \vz_t)$,
history is exponentially decayed, favoring more recent context.

The gating mechanism shows another benefit of \model:
it would be otherwise more difficult to 
build similar techniques into the softmax attention,
where there is no clear sense of ``recurrence'' (Appendix~\ref{sec:gated_softmax}).
It proves useful in our language modeling experiments~(\S\ref{sec:lm}).

\subsection{Discussion}\label{sec:discussion}
\textbf{On query and key norms, and learned random feature variance.}
Eq.~\ref{eq:approx_softmax} assumes both the query and keys
are of norm-1.
It therefore approximates a softmax attention that normalizes
the queries and keys before multiplying them,
and then scales the logits by dividing them by $\sigma^2$.
Empirically, this normalization step
scales down the logits~\citep{vaswani2017attention}
and enforces that $-1\leq\vq^\top\vk\leq1$.
In consequence, the softmax outputs would be ``flattened''
if not for $\sigma$,
which can be set \emph{a priori} as a hyperparameter~\interalia{yu2016orf,avron2017faster,sun2019random}.
Here we instead learn it from data with the reparameterization trick~\citep{kingma2014vae}:
\begin{align}\label{eq:reparam}
    \widetilde{\vw}_i\sim \gN(\vzero, \mI_d), \quad 
    \vw_i = \boldsymbol{\sigma}\circ\widetilde{\vw}_i.
\end{align}
$\mI_d$ is the $d\times d$ identity matrix,
and $\circ$ denotes elementwise product between vectors.
$d$-dimensional vector $\boldsymbol{\sigma}$ is learned,
\rev{but random vectors $\widetilde{\vw}_i$ are \emph{not}.}\footnote{
 This departs from Eq.~\ref{eq:bochner}
by lifting the isotropic assumption imposed on the Gaussian distribution:
note the difference between the vector $\boldsymbol{\sigma}$
in Eq.~\ref{eq:reparam} and the scalar $\sigma$ in Eq.~\ref{eq:approx_gaussian_kernel}.
We find this improves the performance in practice (\S\ref{sec:experiments}),
even though the same result in Theorem~\ref{thm:bochner} may not directly apply.
} 

This norm-1 constraint is never mandatory.
Rather, we employ it for notation clarity and easier implementation.
In preliminary experiments we find
it has little impact on the performance
when $\sigma$ is set properly 
or learned from data.
Eq.~\ref{eq:approx_softmax_norm} in Appendix~\ref{sec:rfa_detail}
presents \model \emph{without} imposing it.

\textbf{Going beyond the Gaussian kernel.}
More broadly, random feature methods can be applied to
a family of shift-invariant kernels, with the Gaussian kernel being one of them.
In the same family, the order-1 arc-cosine kernel~\citep{cho2009kernel}
can be approximated with feature map:
$\boldsymbol{\phi}_{\arccos}(\vx) = \sqrt{1/D}[\relu(\vw_1\cdot\vx),\dots,\relu(\vw_D\cdot\vx)]^\top$~\citep{alber2017empirical}.\footnote{
Apart from replacing the sinusoid functions with $\relu$,
it constructs $\vw_i$ in the same way as Eq.~\ref{eq:reparam}.
}
In our experiments, the Gaussian and arc-cosine variants achieve similar performance. 
This supplements the exploration of alternatives to softmax in attention~\citep{tsai2019transformer,gao2019kernel}.

\textbf{Relations to prior work.}
\citet{katharopoulos20transformers} inspire
the causal attention variant of \model.
They use a feature map based on the  exponential linear unit activation~\citep{clevert2016fast}:
$\operatorname{elu}(\boldsymbol{\cdot}) + 1$.
It significantly \emph{underperforms} both the baseline and \model
in our controlled experiments, showing the importance
of a properly-chosen feature map.
Random feature approximation of attention
is also explored by a concurrent work \citep{choromanski2020masked},
with applications in masked language modeling for proteins.
\rev{They propose positive random features to approximate softmax, 
aiming for a lower variance in critical regions.
\model instead normalizes the queries and keys before random projection to reduce variance.
}
Going beyond both, \model establishes the benefits
of random feature methods as a more universal substitute 
for softmax across all attention variants,
facilitating its applications in, e.g., sequence-to-sequence learning.

There are interesting connections between
gated \model and fast weights \interalia{schmidhuber1992learning,schmidhuber1993reducing,ba2016using,thomas2018differentiable}.
Emphasizing recent patterns,
they learn a temporal memory to store history
similarly to Eqs.~\ref{eq:gate_rfa}.
The main difference is that 
\model additionally 
normalizes the output using $\boldsymbol{\phi}(\vq_t)\cdot\vz$ as in Eq.~\ref{eq:linear_causal},
a by-product of approximating softmax's partition function.
It is intriguing to study the role of this
normalization term, which we leave to future work.

\subsection{Complexity Analysis}\label{sec:complexity}

\textbf{Time.}
Scaling linearly in the sequence lengths, 
\rev{\model needs less computation (in terms of number of operations) for long sequences.}
\rev{This implies speedup wherever the quadratic-time softmax attention
\emph{cannot} be fully-parallelized across time steps.}
More specifically:
\begin{compactitem}
\item Significant speedup can be expected in autoregressive \emph{decoding},
both conditional (e.g., machine translation) 
and unconditional (e.g., sampling from a language model).
For example, 1.9$\times$ speedup is achieved
in our machine translation experiments~(\S\ref{sec:mt});
and more for longer sequences (e.g.,
12$\times$ for 2,048-length ones;~\S\ref{sec:analysis}).
\item Some applications (e.g., language modeling, text classification)
reveal inputs to the model in full.\footnote{A causal masking is usually used to prevent the model from accessing future tokens in language models.}
When there are enough threads to parallelize softmax attention across time steps,
hardly any speedup from \model can be achieved;
when there are not, typically for very long sequences ($>$1,000), substantial speed gain is possible.
For example, \model does \emph{not} achieve any speedup
when working with 512-length context~(\S\ref{sec:lm}),
but achieves a $5.3\times$ speedup with 4,000-length context~(\S\ref{sec:lra}).
\end{compactitem}
\textbf{Memory.}
Asymptotically, \model has a better memory efficiency 
than its softmax counterpart (linear vs.~quadratic).
To reach a more practical conclusion,
we include in our analysis the cost of the feature maps.  
$\boldsymbol{\phi}$'s memory overhead largely depends on its size $D$.
For example, let's consider the cross attention of a decoder.
\model uses $\gO(4D + 2Dd)$ space to store 
$\boldsymbol{\phi}(\vq_t)$,
\rev{$\sum_i \boldsymbol{\phi}(\vk_i)\otimes\vv_i$,
and $\sum_i\boldsymbol{\phi}(\vk_i)$}
(\rev{Eq.~\ref{eq:approx_softmax}}; line~\ref{algo:line:cum_s} of Algo.~\ref{algo:rfa_cross}).\footnote{
\rev{\model \emph{never} constructs the $M\times2D\times d$ tensor 
$[\boldsymbol{\phi}(\vk_i)\otimes\vv_i]_i$,
but sequentially processes the sequence.
}
}
In contrast, 
softmax cross attention 
stores the encoder outputs with  
$\gO(Md)$ memory, with $M$ being 
the source length.
In this case \model has a lower memory overhead
when $2D \ll M$.
Typically $D$ should be no less than $d$ in order for reasonable approximation~\citep{yu2016orf};
In a transformer model, $d$ is the size of an attention head,
which is usually around 64 or 128~\citep{vaswani2017attention,ott2018scaling}.
This  suggests that 
\model can achieve significant memory saving with longer sequences,
which is supported by our empirical analysis in \S\ref{sec:analysis}.
Further, using moderate sized feature maps is also desirable, 
so that its overhead does not overshadow the time and memory \model saves.
We experiment with $D$ at $d$ and $2d$;
the benefit of using $D>2d$ is marginal.

Appendix~\ref{sec:app:complexity}
discusses the time and space complexity in more detail,
\rev{and Appendix~\ref{sec:random_feature_size} studies
the effect of random feature size on performance.
}
\section{Experiments}\label{sec:experiments}
We evaluate \model on language modeling, machine translation, \rev{and long text classification}.

\subsection{Language Modeling}\label{sec:lm}

\textbf{Setting.}
We experiment with WikiText-103~\citep{merity2016pointer}.
It is based on  English Wikipedia.
Table~\ref{tab:data} in Appendix~\ref{sec:experimental_details} summarizes some of its statistics.
We compare the following models:
\begin{compactitem}
\item \base is our implementation of the strong transformer-based language model by~\citet{baevski2018adaptive}. 
\item \model builds on \base, but replaces the softmax attention 
with random feature attention.  
We experiment with both Gaussian and arc-cosine kernel variants.
\item \modelgate additionally learns a sigmoid gate on top of \model (\S\ref{sec:gate}). It also has a Gaussian kernel variant and a arc-cosine kernel one.\footnote{ 
This gating technique is specific to \model variants, 
in the sense that it is less intuitive to apply it in \base.}
\item $\boldsymbol{\phi}_{\operatorname{elu}}$ is a baseline to \model.
Instead of the random feature methods it uses the 
$\operatorname{elu}(\boldsymbol{\cdot}) + 1$ feature map, 
as in \citet{katharopoulos20transformers}. 
\end{compactitem}
To ensure fair comparisons, we use comparable implementations, tuning, and
training procedure.
All models use a 512 block size during both training and evaluation, i.e., 
they read as input a segment of 512 consecutive tokens, \emph{without}
access to the context from previous mini-batches. 
\model variants use 64-dimensional random feature maps.
We experiment with two model size settings,
\term{small} (around 38M parameters) and \term{big} (around 242M parameters);
they are described in Appendix~\ref{sec:lm_details}
along with other implementation details.

\begin{table*}[th]
\centering
\begin{tabular}{@{} l cc m{0.01em} cc@{}}
\toprule[.1em]

& \multicolumn{2}{c}{\textbf{Small}}
&& \multicolumn{2}{c}{\textbf{Big}}\\
\cmidrule(lr){2-3}
\cmidrule(lr){5-6}
\textbf{Model} 
& \textbf{Dev.} & \textbf{Test}
&& \textbf{Dev.} & \textbf{Test}\\

\midrule[.1em]
\base 
&33.0 & 34.5
&& 24.5 & 26.2\\

\midrule
$\boldsymbol{\phi}_{\operatorname{elu}}$~\citep{katharopoulos20transformers}

&38.4 & 40.1
& & 28.7  & 30.2\\

\midrule

\model-Gaussian 
&33.6 & 35.7
&& 25.8 &  27.5\\ 

\model-$\arccos$
&36.0 & 37.7
&& 26.4 &  28.1 \\

\midrule

\modelgate-Gaussian  
& \textbf{31.3} &  \textbf{32.7}
&& \textbf{23.2} &  \textbf{25.0}\\

\modelgate-$\arccos$
& \textbf{32.8} &  \textbf{34.0}
&& 24.8 &  26.3\\ 

\midrule

\modelgate-Gaussian-Stateful 
& \textbf{29.4} & \textbf{30.5}
&& \textbf{22.0} & \textbf{23.5}\\

\bottomrule[.1em]
\end{tabular}
\caption{\label{tab:lm_results}Language model perplexity ({lower is better}) on the   WikiText-103 
development and test sets. Bolded numbers outperform \base.
} 
\end{table*}

\textbf{Results.}
Table~\ref{tab:lm_results} compares the models'
performance in perplexity 
on WikiText-103 development and test data.
Both kernel variants of \model, \emph{without} gating,
outperform $\boldsymbol{\phi}_{\operatorname{elu}}$
by more than 2.4 and 2.1 test perplexity for the small and big model respectively, 
confirming the benefits from using random feature approximation.\footnote{
    All models are trained for 150K steps;
    this could be part of the reason behind 
    the suboptimal performance of $\boldsymbol{\phi}_{\operatorname{elu}}$: 
    it may need 3 times more gradient updates
    to reach similar performance to the softmax attention baseline~\citep{katharopoulos20transformers}.
}
Yet both \emph{underperform} \base, with
\model-Gaussian having a smaller gap.
Comparing \model against its gated variants, 
a more than 1.8 perplexity improvement can be attributed
to the gating mechanism;
and the gap is larger for small models.
Notably, \modelgate-Gaussian outperforms \base
under both size settings by at least 1.2 perplexity.
In general, \model models with Gaussian feature maps outperform
their arc-cosine counterparts.\footnote{
    We observe that \model Gaussian variants
    are more stable and easier to train than the arc-cosine ones as well as $\boldsymbol{\phi}_{\operatorname{elu}}$.
    We conjecture that this is because the outputs of the Gaussian feature maps have an $\ell_2$-norm of 1,
    which can help stabilize training. 
    To see why, $\sin^2(x) + \cos^2(x)=\cos(x-x)=1$.
}
From the analysis in \S\ref{sec:complexity} we would \emph{not} expect 
speedup by \model models, nor do we see any in the experiments.\footnote{
In fact, \model \emph{trains} around 15\% slower than \base due to the additional overhead from the feature maps. 
}

Closing this section, we explore a ``stateful'' variant of  \modelgate-Gaussian. 
It passes the last hidden state $(\mS_t, \vz_t)$
to the next mini-batch during both training and evaluation, 
a technique commonly used in RNN language models~\citep{merity2018reg}.
This is a consequence of \model's RNN-style computation,
and is less straightforward to be applicable in the vanilla transformer models.\footnote{Some transformer models use a 
text segment from the previous mini-batch as a prefix
\citep{baevski2018adaptive,dai2019}. Unlike \model, this gives the model access to only a limited amount of context, and significantly increases the memory overhead. 
}
From the last row of Table~\ref{tab:lm_results} 
we see that this brings a
more than 1.5 test perplexity improvement.

\subsection{Machine Translation}\label{sec:mt}
\textbf{Datasets.} We experiment with three standard machine translation datasets.

\begin{itemize}[nosep,leftmargin=1em,labelwidth=*,align=left]
\item WMT14 EN-DE and EN-FR~\citep{bojar2014wmt}.
Our data split and preprocessing follow those of \citet{vaswani2017attention}.
We share the source and target vocabularies within each language pair, 
with 32,768 byte pair encoding types (BPE;~\citealp{sennrich2016bpe}).
\item IWSLT14 DE-EN~\citep{cettolo2014report} is based on TED talks.
The preprocessing follows~\citet{edunov2018classical}.
Separate vocabularies of 9K/7K BPE types are used for the source and target.
\end{itemize}
Table~\ref{tab:data} in Appendix~\ref{sec:experimental_details} summarizes some statistics of the datasets.

\textbf{Setting.}
We compare the \model variants described in~\S\ref{sec:lm}.
They build on a \base model that is our implementation of the base-sized
transformer~\citep{vaswani2017attention}.
All \model models apply random feature attention 
in decoder cross and causal attention, but use softmax attention in encoders.
This setting yields the greatest decoding time and memory savings (\S\ref{sec:complexity}).
We use 128/64 for $D$ in cross/causal attention.
\modelgate learns sigmoid gates in the decoder causal attention.
The $\boldsymbol{\phi}_{\operatorname{elu}}$ baseline 
uses the same setting and applies
feature map in both decoder cross and causal attention,
but \emph{not} in the encoders.
Further details are described in Appendix~\ref{sec:mt_details}.

\begin{table*}[th]
\centering
\begin{tabular}{@{} l cc  m{0.001em}   c c@{}}
\toprule[.1em]
&
\multicolumn{2}{c}{\textbf{WMT14}} &&  
\textbf{IWSLT14}\\
\cmidrule(lr){2-3}
\cmidrule(lr){5-5}
\textbf{Model} 
& \textbf{EN-DE}  & \textbf{EN-FR}
&& \textbf{DE-EN} & \textbf{Speed}\\

\midrule[.1em]
\base & 28.1 & 39.0 
&& 34.6 & $1.0\times$\\ 

\midrule

$\boldsymbol{\phi}_{\operatorname{elu}}$~\citep{katharopoulos20transformers}
& 21.3 & 34.0 && 29.9 & $2.0\times$\\

\midrule
\model-Gaussian & 28.0 & 39.2
&& 34.5& $1.8\times$\\ 
\model-$\arccos$ & 28.1 & 38.9 
&& 34.4 & $1.9\times$\\ 

\midrule
\modelgate-Gaussian & 28.1 & 39.0
&& 34.6 & $1.8\times$\\
\modelgate-$\arccos$  & 28.2 & 39.2 
&& 34.4 & $1.9\times$ \\ 

\bottomrule[.1em]
\end{tabular}
\caption{\label{tb:mt_results}Machine translation test set BLEU. The decoding speed (last column)
is relative to \base.
All models are tested on a single TPU v2 accelerator, with  batch size  32.} 
\end{table*}

\textbf{Results.}
Table~\ref{tb:mt_results} compares the models' test set BLEU
on three machine translation datasets.
Overall both Gaussian and arc-cosine variants of \model
achieve similar performance to \base on all three datasets,
significantly outperforming \citet{katharopoulos20transformers}.
Differently from the trends in the language modeling experiments,
here the gating mechanism does not lead to substantial gains.
Notably, all \model variants decode more than $1.8\times$
faster than \base.

\rev{\subsection{Long Text Classification}}\label{sec:lra}
\rev{
We further evaluate  \model's accuracy and efficiency when used as text encoders on three NLP tasks from the recently proposed 
Long Range Arena benchmark~\citep{tay2020long}, 
designed to evaluate efficient Transformer variants on tasks that require processing long sequences.\footnote{
\url{https://github.com/google-research/long-range-arena}
}}

\rev{
\textbf{Experimental setting and datasets.}
We compare \model against baselines on the following datasets:
\begin{compactitem}
\item ListOps (\textbf{LO};~\citealp{nangia2018listops}) aims to diagnose the capability of modelling hierarchically structured data.
Given a sequence of operations on single-digit integers, 
the model predicts the solution, also a single-digit integer.
It is formulated as a 10-way classification.
We follow \citet{tay2020long} and consider sequences with 500--2,000 symbols.
\item Character-level text classification with the \textbf{IMDb} movie review dataset~\citep{maas2011learning}. 
This is a binary sentiment classification task.
\item Character-level document retrieval with the ACL Anthology Network (\textbf{AAN};~\citealp{radev2009aan}) dataset.
The model classifies whether there is a citation between a pair of papers.
\end{compactitem}
To ensure fair comparisons, 
we implement \model on top of the transformer baseline by \citet{tay2020long},
and closely follow their preprocessing, data split, model size, and training procedure.
Speed and memory are evaluated on the IMDb dataset.
For our \model model, we use $D=64$ for the IMDb dataset, and $D=128$ for others.
We refer the readers to \citet{tay2020long} for further details.
}

\rev{
\textbf{Results.}
From Table~\ref{tab:lra} we can see that 
\model outperforms the transformer
baseline on two out of the three datasets,
achieving the best performance on IMDb with 66\% accuracy.
Averaging across three datasets, 
\model outperforms the transformer by 0.3\% accuracy,
second only to \citet{zaheer2020big} with a 0.1\% accuracy gap.
In terms of time and memory efficiency, 
\model is among the strongest.
\model speeds up over the transformer by $1.1$--$5.3\times$, varying by sequence length.
Importantly, compared to the only two baselines that perform comparably to the baseline transformer model \citep{tay2020synthesizer,zaheer2020big}, \model has a clear advantage in both speed and memory efficiency, and is the only model that is competitive in both accuracy and efficiency.
}

\begin{table*}[t]
\centering
\small
\begin{tabular}{@{} 
l
@{\hspace{12pt}}
c@{\hspace{8pt}}c@{\hspace{8pt}}c@{\hspace{8pt}}c 
@{} m{14pt} @{}
c@{\hspace{8pt}}c@{\hspace{8pt}}c@{\hspace{8pt}}c  
@{} m{14pt} @{}
c@{\hspace{8pt}}c@{\hspace{8pt}}c@{\hspace{8pt}}c 
@{}}
\toprule[.1em]

& \multicolumn{4}{c}{\textbf{Accuracy}}
&& \multicolumn{4}{c}{\textbf{Speed}}
&& \multicolumn{4}{c}{\textbf{Memory}}\\
\cmidrule(lr){2-5}
\cmidrule(lr){7-10}
\cmidrule(lr){12-15}

\textbf{Model}
& \textbf{LO} & \textbf{IMDb}
& \textbf{AAN} & \textbf{Avg.}
&& \textbf{1K} & \textbf{2K}
& \textbf{3K} & \textbf{4K}
&& \textbf{1K} & \textbf{2K}
& \textbf{3K} & \textbf{4K}\\

\midrule[.1em]
Transformer 
& 36.4 & 64.3 & 57.5 
& 52.7 
&& 1.0 & 1.0
& 1.0 & 1.0
&& 1.00 & 1.00
& 1.00 & 1.00\\
\midrule

\citet{wang2020linformer} 
& 35.7 & 53.9 & 52.3 
& 47.3
 && 1.2 & 1.9
& 3.7 & 5.5
&& \textbf{0.44} & \textbf{0.21}
& 0.18 & \textbf{0.10}\\

\citet{kitaev2020reformer} 
& \underline{\textbf{37.3}} & 56.1 & 53.4 
& 48.9
&& 0.5 & 0.4
& 0.7 & 0.8
&& 0.56 & 0.37
& 0.28 & 0.24\\

\citet{tay20sparse} 
& 17.1 & 63.6 & \underline{\textbf{59.6}} 
& 46.8
&& 1.1 & 1.6
& 2.9 & 3.8
&& 0.55 & 0.31
& 0.20 & 0.16\\

\citet{tay2020synthesizer}
& \underline{37.0} & 61.7 & 54.7
& 51.1
&& 1.1 & 1.2
& 2.9 & 1.4
&& 0.76 & 0.75
& 0.74 & 0.74\\

\citet{zaheer2020big}
& 36.0 & 64.0 & \underline{59.3}
& \underline{\textbf{53.1}}
&& 0.9 & 0.8
& 1.2 & 1.1
&& 0.90 & 0.56
& 0.40 & 0.30\\

\citet{katharopoulos20transformers} 
& 16.1 & \underline{65.9} & 53.1 
& 45.0
&& 1.1 & \textbf{1.9}
& 3.7 & 5.6
&& 0.44 & 0.22
& \textbf{0.14} & 0.11\\

\citet{choromanski2020masked}
& 18.0 & \underline{65.4} & 53.8 
& 45.7
&& \textbf{1.2} & \textbf{1.9}
& \textbf{3.8} & \textbf{5.7}
&& 0.44 & 0.22
& 0.15 & 0.11\\

\midrule
RFA-Gaussian (This work)
& \underline{36.8} & \underline{\textbf{66.0}} & 56.1
& \underline{53.0}
&& 1.1 & 1.7
& 3.4 & 5.3
&& 0.53 & 0.30
& 0.21 & 0.16\\

\bottomrule[.1em]
\end{tabular}

\caption{\label{tab:lra}
Accuracy  (higher is better) of different models on LO, IMDb, and AAN,
along with their
speed (higher is better) and peak memory consumption (lower is better) 
varying sequence lengths (1--4K). 
Speed and memory are evaluated on the IMDb dataset and relative to the transformer's.
Bold font indicates the best performance in each column,
and underlined numbers outperform the transformer in accuracy.
Transformer's and previous works' numbers are due to \citet{tay2020long}.}
\vspace{-0.5cm}
\end{table*}

\section{Analysis}\label{sec:analysis}

\begin{figure}[t]
\centering
\begin{subfigure}[t]{0.49\textwidth}
\includegraphics[trim={.5cm 0cm 2.5cm .0cm},clip,width=\textwidth]{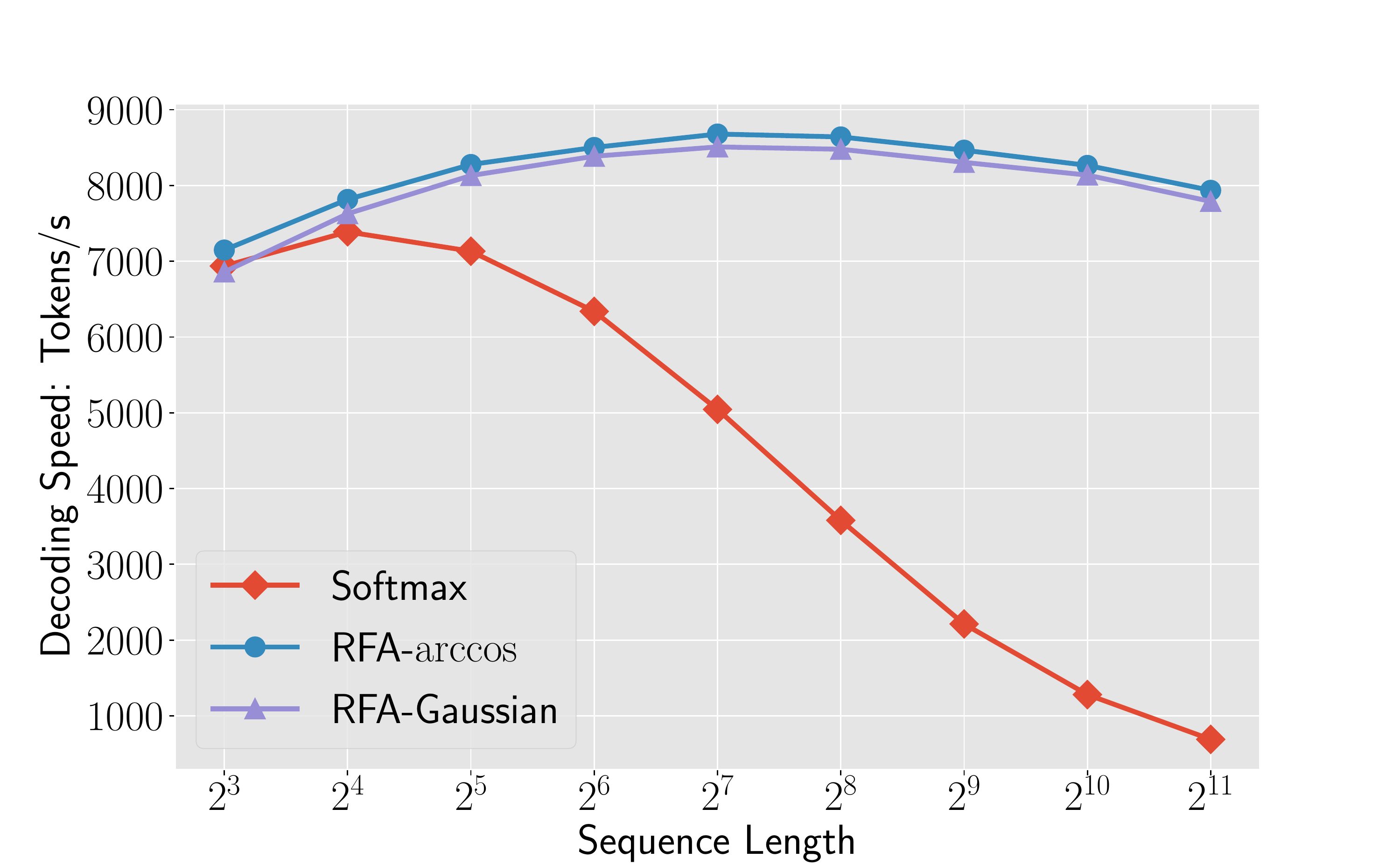}
\caption{Speed vs. lengths.}
\end{subfigure}
\begin{subfigure}[t]{0.49\textwidth}
\includegraphics[trim={.5cm 0cm 2.5cm .0cm},clip,width=\textwidth]{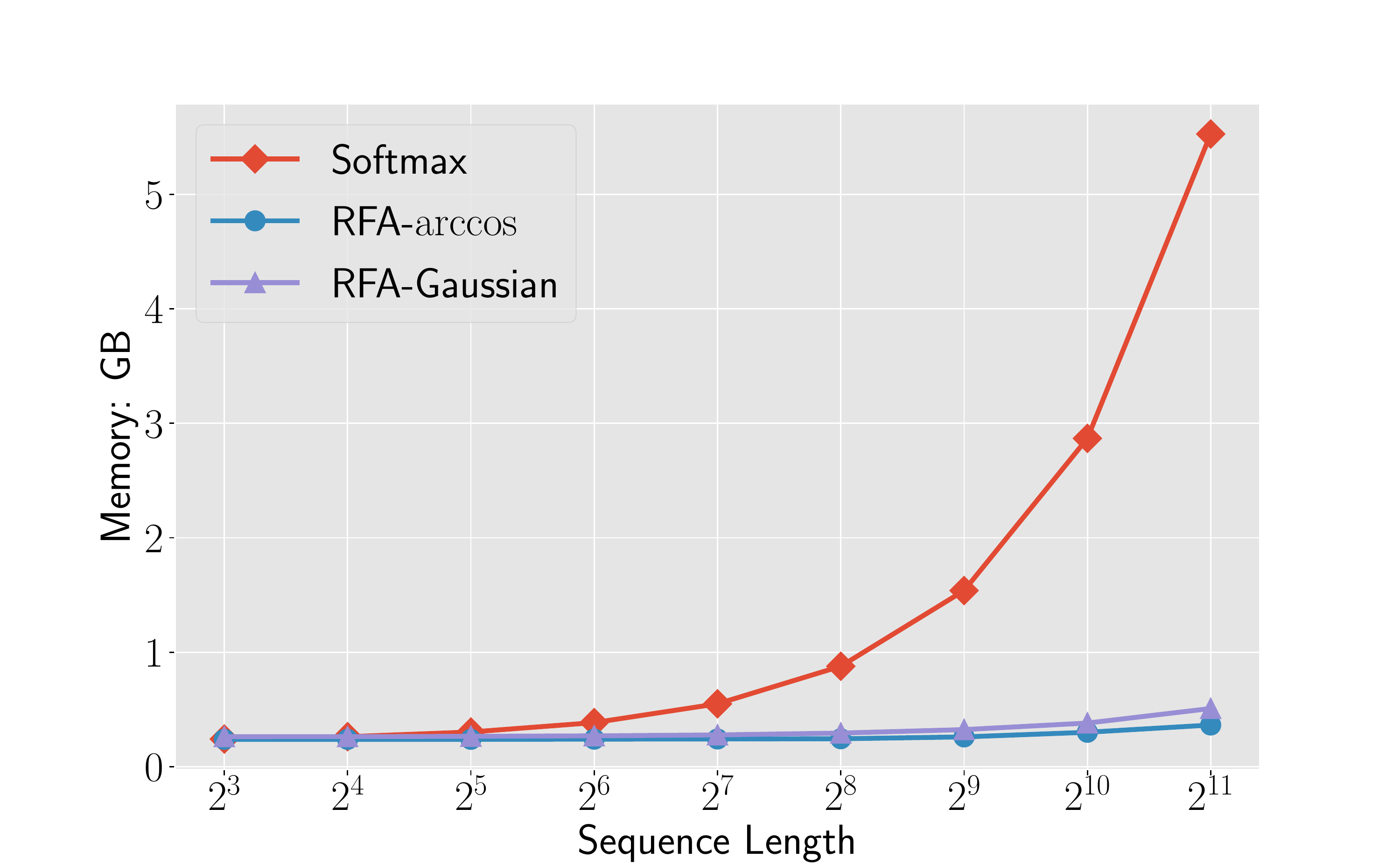}
\caption{\label{fig:length_memory}Memory vs. lengths.}
\end{subfigure}
\caption{Conditional decoding speed (left) and memory overhead (right) varying the output lengths. All models are tested on a single TPU v2 accelerator, with greedy decoding and batch size 16.}\label{fig:length_conditional}
\vspace{-.5cm}
\end{figure}

\textbf{Decoding time and memory varying by sequence length.}
\S\ref{sec:complexity} shows that \model can potentially 
achieve more significant speedup and memory saving for longer sequences, which we now explore.

We use a simulation conditional generation experiment on to compare
\model's sequence-to-sequence decoding speed and memory overhead
against the baseline's.
Here we assume the input and output sequences are of the same length.
The compared models are of the same size as those described in \S\ref{sec:mt},
with 6-layer encoders and decoders.
Other hyperparameters are summarized in Appendix~\ref{sec:mt_details}.
All models are tested using greedy decoding with the same batch size of 16, on a TPU v2 accelerator.

From Figures~\ref{fig:length_conditional} (a) and (b) we observe clear trends.
Varying the lengths,
both \model variants achieve consistent decoding speed with
nearly-constant memory overhead.
In contrast, the baseline decodes slower for longer sequences, taking 
an increasing amount of memory.
Notably, for 2,048-length sequences,
\model decodes around 12$\times$ faster than the baseline while using less than 10\%
of the memory.
\model-$\arccos$ slightly outperforms \model-Gaussian in terms of speed and memory efficiency.
This is because when using the same $D$ (as we do here),
the $\boldsymbol{\phi}_{\arccos}$ is half the size of $\boldsymbol{\phi}_{\text{Gaussian}}$.
These results suggest that \model can be particularly useful
in sequence-to-sequence tasks with longer sequences,
e.g., document-level machine translation~\citep{miculicich2018document}.

Figure~\ref{fig:length_unconditional} in Appendix~\ref{sec:more_length}
compares the speed and memory consumption
in \emph{unconditional} decoding (e.g., sampling from a language model).
The overall trends are similar to those in Figure~\ref{fig:length_conditional}.

\textbf{Notes on decoding speed.}
With a lower memory overhead, \model can use
a larger batch size than the baseline.
As noted by~\citet{katharopoulos20transformers} and \citet{kasai2020deep},
if we had used mini-batches as large as the hardware allows,
\model could have achieved a more significant speed gain.
Nonetheless, we control for batch size
even though it is not the most favorable setting for \model, 
since the conclusion
translates better to common applications
where one generates a single sequence at a time (e.g., instantaneous machine translation).
For the softmax attention baseline, we follow \citet{ott2018scaling} and cache previously computed query/key/value representations,
which significantly improves its decoding speed (over not caching).

\textbf{Further analysis results.}
\model achieves comparable performance to softmax attention.
Appendix~\ref{sec:train_eval} empirically shows that 
this \emph{cannot} be attributed to 
\model learning a good approximation to softmax:
when we train with one attention but evaluate with the other,
the performance is hardly better than randomly-initialized untrained models.
Yet, an \model model
initialized from a pretrained softmax transformer 
achieves decent training loss after a 
moderate amount of finetuning steps~(Appendix~\ref{sec:finetune}).
This suggests some potential applications,
e.g., transferring knowledge
from a pretrained transformer
(e.g., GPT-3;~\citealp{brown2020language})
to an \model model that is more efficient to sample from. 
\section{Related Work}
One common motivation across the following studies, 
that is shared by this work and the research we have already discussed, is to scale  
transformers to long sequences. 
Note that there are plenty orthogonal choices for improving efficiency such as weight sharing \citep{dehghani2018universal}, quantization \citep{shen2020}, knowledge distillation \citep{Sanh2019DistilBERTAD}, and adapters \citep{pmlr-v97-houlsby19a}. 
For a detailed overview we refer the reader to \cite{tay2020}.
  
\textbf{Sparse attention patterns.}
The idea behind these methods 
is to limit
the reception field of attention computation.
It motivates earlier attempts in
improving attention's efficiency,
and still receives lots of interest.
The sparse patterns can be set \textit{a priori}~\interalia{liu2018generating,qiu2019blockwise,ho2020axial,you2020hard}
or learned from data~\interalia{sukhbaatar2019adaptive,roy2020efficient}.
For most of these approaches,
it is yet to be empirically verified that 
they are suitable for large-scale sequence-to-sequence learning;
few of them have recorded decoding speed benefits.

\textbf{Compressed context.}
\citet{wang2020linformer} compress the 
context along the timesteps
so that the effective sequence length for attention
computation is reduced.
Another line of work aims to store
past context into a memory module with limited size~\interalia{lee2019set,ainslie2020etc,Rae2020Compressive},
so that accessing longer history only moderately increases the overhead.
Reminiscent of RNN language models,
\model attends beyond a fixed context window
through a stateful computation, \emph{without} increasing
time or memory overhead.

\section{Conclusion}\label{sec:conclusion}
We presented random feature attention (\model).
It views the softmax attention through
the lens of kernel methods,
and approximates it with  random feature methods.
With an optional gating mechanism,
\model provides a straightforward way
of learning with recency bias.
\model's time and space complexity 
is linear in the sequence length.
We use \model as a drop-in substitute for softmax attention 
in transformer models.
On language modeling, machine translation, 
and long text classification benchmarks,
\model achieves comparable or better performance
than strong baselines.
In the machine translation experiment,
\model decodes twice as fast. 
Further time and memory efficiency
improvements can be achieved for longer sequences.

\section*{Acknowledgments}
	We would like to thank 
	Phil Blunsom,
	Chris Dyer,
	Nando de Freitas,
	Jungo Kasai,
	Adhiguna Kuncoro,
	Dianqi Li,
	Ofir Press,
	Lianhui Qin, 
	Swabha Swayamdipta,
	Sam Thomson,
	the language team at DeepMind
	and the ARK group at the University of Washington
	for their helpful feedback.
	We also thank Tay Yi
	for helping run the Long Range Arena experiments,
	Richard Tanburn
	for the advice on implementations,
    and the anonymous reviewers
	for their thoughtful comments.
	This work was supported in part by NSF grant 1562364 and a Google Fellowship. Nikolaos Pappas was supported by the Swiss National Science Foundation under grant number  P400P2\_183911 ``UNISON.''

\bibliography{iclr2021_conference}

\begin{thebibliography}{78}
\providecommand{\natexlab}[1]{#1}
\providecommand{\url}[1]{\texttt{#1}}
\expandafter\ifx\csname urlstyle\endcsname\relax
  \providecommand{\doi}[1]{doi: #1}\else
  \providecommand{\doi}{doi: \begingroup \urlstyle{rm}\Url}\fi

\bibitem[Ainslie et~al.(2020)Ainslie, Ontanon, Alberti, Cvicek, Fisher, Pham,
  Ravula, Sanghai, Wang, and Yang]{ainslie2020etc}
Joshua Ainslie, Santiago Ontanon, Chris Alberti, Vaclav Cvicek, Zachary Fisher,
  Philip Pham, Anirudh Ravula, Sumit Sanghai, Qifan Wang, and Li~Yang.
\newblock {ETC}: Encoding long and structured inputs in transformers.
\newblock In \emph{Proc. of EMNLP}, 2020.

\bibitem[Alber et~al.(2017)Alber, Kindermans, Sch\"{u}tt, M\"{u}ller, and
  Sha]{alber2017empirical}
Maximilian Alber, Pieter-Jan Kindermans, Kristof Sch\"{u}tt, Klaus-Robert
  M\"{u}ller, and Fei Sha.
\newblock An empirical study on the properties of random bases for kernel
  methods.
\newblock In \emph{Proc. of NeurIPS}, 2017.

\bibitem[Avron et~al.(2016)Avron, Sindhwani, Yang, and Mahoney]{avron2016qmc}
Haim Avron, Vikas Sindhwani, Jiyan Yang, and Michael~W. Mahoney.
\newblock Quasi-{Monte Carlo} feature maps for shift-invariant kernels.
\newblock \emph{Journal of Machine Learning Research}, 17\penalty0
  (120):\penalty0 1--38, 2016.

\bibitem[Avron et~al.(2017)Avron, Clarkson, and Woodruff]{avron2017faster}
Haim Avron, L.~Kenneth Clarkson, and P.~David~and Woodruff.
\newblock Faster kernel ridge regression using sketching and preconditioning.
\newblock \emph{SIAM J. Matrix Analysis Applications}, 2017.

\bibitem[Ba et~al.(2016)Ba, Hinton, Mnih, Leibo, and Ionescu]{ba2016using}
Jimmy Ba, Geoffrey~E Hinton, Volodymyr Mnih, Joel~Z Leibo, and Catalin Ionescu.
\newblock Using fast weights to attend to the recent past.
\newblock In \emph{Proc. of NeurIPS}, 2016.

\bibitem[Baevski \& Auli(2019)Baevski and Auli]{baevski2018adaptive}
Alexei Baevski and Michael Auli.
\newblock Adaptive input representations for neural language modeling.
\newblock In \emph{Proc. of ICLR}, 2019.

\bibitem[Bahdanau et~al.(2015)Bahdanau, Cho, and Bengio]{bahdanau2015attention}
Dzmitry Bahdanau, Kyunghyun Cho, and Yoshua Bengio.
\newblock Neural machine translation by jointly learning to align and
  translate.
\newblock In \emph{Proc. of ICLR}, 2015.

\bibitem[Beltagy et~al.(2020)Beltagy, Peters, and Cohan]{beltagy2020longformer}
Iz~Beltagy, Matthew~E. Peters, and Arman Cohan.
\newblock Longformer: The long-document transformer.
\newblock \emph{arXiv: 2004.05150}, 2020.

\bibitem[Bochner(1955)]{bochner1955harmonic}
S.~Bochner.
\newblock \emph{Harmonic Analysis and the Theory of Probability}.
\newblock University of California Press, 1955.

\bibitem[Bojar et~al.(2014)Bojar, Buck, Federmann, Haddow, Koehn, Leveling,
  Monz, Pecina, Post, Saint-Amand, Soricut, Specia, and Tamchyna]{bojar2014wmt}
Ond{\v{r}}ej Bojar, Christian Buck, Christian Federmann, Barry Haddow, Philipp
  Koehn, Johannes Leveling, Christof Monz, Pavel Pecina, Matt Post, Herve
  Saint-Amand, Radu Soricut, Lucia Specia, and Ale{\v{s}} Tamchyna.
\newblock Findings of the 2014 workshop on statistical machine translation.
\newblock In \emph{Proc. of WMT}, 2014.

\bibitem[Brown et~al.(2020)Brown, Mann, Ryder, Subbiah, Kaplan, Dhariwal,
  Neelakantan, Shyam, Sastry, Askell, Agarwal, Herbert-Voss, Krueger, Henighan,
  Child, Ramesh, Ziegler, Wu, Winter, Hesse, Chen, Sigler, Litwin, Gray, Chess,
  Clark, Berner, McCandlish, Radford, Sutskever, and Amodei]{brown2020language}
Tom~B. Brown, Benjamin Mann, Nick Ryder, Melanie Subbiah, Jared Kaplan,
  Prafulla Dhariwal, Arvind Neelakantan, Pranav Shyam, Girish Sastry, Amanda
  Askell, Sandhini Agarwal, Ariel Herbert-Voss, Gretchen Krueger, Tom Henighan,
  Rewon Child, Aditya Ramesh, Daniel~M. Ziegler, Jeffrey Wu, Clemens Winter,
  Christopher Hesse, Mark Chen, Eric Sigler, Mateusz Litwin, Scott Gray,
  Benjamin Chess, Jack Clark, Christopher Berner, Sam McCandlish, Alec Radford,
  Ilya Sutskever, and Dario Amodei.
\newblock Language models are few-shot learners.
\newblock \emph{arXiv: 2005.14165}, 2020.

\bibitem[Cettolo et~al.(2014)Cettolo, Niehues, Stüker, Bentivogli, and
  Federico]{cettolo2014report}
Mauro Cettolo, Jan Niehues, Sebastian Stüker, Luisa Bentivogli, and Marcello
  Federico.
\newblock Report on the 11th {IWSLT} evaluation campaign.
\newblock In \emph{Proc. of IWSLT}, 2014.

\bibitem[Chen et~al.(2019)Chen, Wang, Utiyama, and Sumita]{chen2019recurrent}
Kehai Chen, Rui Wang, Masao Utiyama, and Eiichiro Sumita.
\newblock Recurrent positional embedding for neural machine translation.
\newblock In \emph{Proc. of EMNLP}, 2019.

\bibitem[Child et~al.(2019)Child, Gray, Radford, and
  Sutskever]{child2019generating}
Rewon Child, Scott Gray, Alec Radford, and Ilya Sutskever.
\newblock Generating long sequences with sparse transformers.
\newblock \emph{arXiv: 1904.10509}, 2019.

\bibitem[Cho et~al.(2014)Cho, van Merri{\"e}nboer, Gulcehre, Bahdanau,
  Bougares, Schwenk, and Bengio]{cho2014gru}
Kyunghyun Cho, Bart van Merri{\"e}nboer, Caglar Gulcehre, Dzmitry Bahdanau,
  Fethi Bougares, Holger Schwenk, and Yoshua Bengio.
\newblock Learning phrase representations using {RNN} encoder{--}decoder for
  statistical machine translation.
\newblock In \emph{Proc. of EMNLP}, 2014.

\bibitem[Cho \& Saul(2009)Cho and Saul]{cho2009kernel}
Youngmin Cho and Lawrence~K. Saul.
\newblock Kernel methods for deep learning.
\newblock In \emph{Proc. of NeurIPS}, 2009.

\bibitem[Choromanski et~al.(2020)Choromanski, Likhosherstov, Dohan, Song, Gane,
  Sarlos, Hawkins, Davis, Mohiuddin, Kaiser, Belanger, Colwell, and
  Weller]{choromanski2020masked}
Krzysztof Choromanski, Valerii Likhosherstov, David Dohan, Xingyou Song,
  Andreea Gane, Tamas Sarlos, Peter Hawkins, Jared Davis, Afroz Mohiuddin,
  Lukasz Kaiser, David Belanger, Lucy Colwell, and Adrian Weller.
\newblock Rethinking attention with performers.
\newblock In \emph{Proc. of ICLR}, 2020.

\bibitem[Clevert et~al.(2016)Clevert, Unterthiner, and
  Hochreiter]{clevert2016fast}
Djork-Arné Clevert, Thomas Unterthiner, and Sepp Hochreiter.
\newblock Fast and accurate deep network learning by exponential linear units
  ({ELUs}).
\newblock In \emph{Proc. of ICLR}, 2016.

\bibitem[Dai et~al.(2019)Dai, Yang, Yang, Carbonell, Le, and
  Salakhutdinov]{dai2019}
Zihang Dai, Zhilin Yang, Yiming Yang, Jaime Carbonell, Quoc Le, and Ruslan
  Salakhutdinov.
\newblock Transformer-{XL}: Attentive language models beyond a fixed-length
  context.
\newblock In \emph{Proc. of ACL}, 2019.

\bibitem[Dehghani et~al.(2019)Dehghani, Gouws, Vinyals, Uszkoreit, and
  Kaiser]{dehghani2018universal}
Mostafa Dehghani, Stephan Gouws, Oriol Vinyals, Jakob Uszkoreit, and Lukasz
  Kaiser.
\newblock Universal transformers.
\newblock In \emph{Proc. of ICLR}, 2019.

\bibitem[Devlin et~al.(2019)Devlin, Chang, Lee, and Toutanova]{delvin2019bert}
Jacob Devlin, Ming-Wei Chang, Kenton Lee, and Kristina Toutanova.
\newblock {BERT}: Pre-training of deep bidirectional transformers for language
  understanding.
\newblock In \emph{Proc. of NAACL}, 2019.

\bibitem[Edunov et~al.(2018)Edunov, Ott, Auli, Grangier, and
  Ranzato]{edunov2018classical}
Sergey Edunov, Myle Ott, Michael Auli, David Grangier, and Marc{'}Aurelio
  Ranzato.
\newblock Classical structured prediction losses for sequence to sequence
  learning.
\newblock In \emph{Proc. of NAACL}, 2018.

\bibitem[Gao et~al.(2019)Gao, Herold, Wang, and Ney]{gao2019kernel}
Yingbo Gao, Christian Herold, Weiyue Wang, and Hermann Ney.
\newblock Exploring kernel functions in the softmax layer for contextual word
  classification.
\newblock In \emph{International Workshop on Spoken Language Translation},
  2019.

\bibitem[Hao et~al.(2019)Hao, Wang, Yang, Wang, Zhang, and Tu]{hao2019modeling}
Jie Hao, Xing Wang, Baosong Yang, Longyue Wang, Jinfeng Zhang, and Zhaopeng Tu.
\newblock Modeling recurrence for transformer.
\newblock In \emph{Proc. of NAACL}, 2019.

\bibitem[Hinton et~al.(2015)Hinton, Vinyals, and Dean]{hinton2015distilling}
Geoffrey Hinton, Oriol Vinyals, and Jeffrey Dean.
\newblock Distilling the knowledge in a neural network.
\newblock In \emph{NeurIPs Deep Learning and Representation Learning Workshop},
  2015.

\bibitem[Ho et~al.(2020)Ho, Kalchbrenner, Weissenborn, and
  Salimans]{ho2020axial}
Jonathan Ho, Nal Kalchbrenner, Dirk Weissenborn, and Tim Salimans.
\newblock Axial attention in multidimensional transformers.
\newblock \emph{arXiv: 1912.12180}, 2020.

\bibitem[Hochreiter \& Schmidhuber(1997)Hochreiter and
  Schmidhuber]{hochreiter1997lstm}
Sepp Hochreiter and J{\"u}rgen Schmidhuber.
\newblock Long short-term memory.
\newblock \emph{Neural Computation}, 9\penalty0 (8):\penalty0 1735--1780, 1997.

\bibitem[Hofmann et~al.(2008)Hofmann, Sch\"{o}lkopf, and
  Smola]{hofmann2008kernel}
Thomas Hofmann, Bernhard Sch\"{o}lkopf, and Alexander~J. Smola.
\newblock Kernel methods in machine learning.
\newblock \emph{Annals of Statistics}, 36\penalty0 (3):\penalty0 1171--1220,
  2008.

\bibitem[Houlsby et~al.(2019)Houlsby, Giurgiu, Jastrzebski, Morrone,
  De~Laroussilhe, Gesmundo, Attariyan, and Gelly]{pmlr-v97-houlsby19a}
Neil Houlsby, Andrei Giurgiu, Stanislaw Jastrzebski, Bruna Morrone, Quentin
  De~Laroussilhe, Andrea Gesmundo, Mona Attariyan, and Sylvain Gelly.
\newblock Parameter-efficient transfer learning for {NLP}.
\newblock In \emph{Proc. of ICML}, 2019.

\bibitem[Kasai et~al.(2021)Kasai, Pappas, Peng, Cross, and
  Smith]{kasai2020deep}
Jungo Kasai, Nikolaos Pappas, Hao Peng, James Cross, and Noah~A. Smith.
\newblock Deep encoder, shallow decoder: Reevaluating the speed-quality
  tradeoff in machine translation.
\newblock In \emph{Proc. of ICLR}, 2021.

\bibitem[Katharopoulos et~al.(2020)Katharopoulos, Vyas, Pappas, and
  Fleuret]{katharopoulos20transformers}
Angelos Katharopoulos, Apoorv Vyas, Nikolaos Pappas, and Francois Fleuret.
\newblock Transformers are rnns: Fast autoregressive transformers with linear
  attention.
\newblock In \emph{Proc. of ICML}, 2020.

\bibitem[Kingma \& Ba(2015)Kingma and Ba]{kingma2014adam}
Diederik Kingma and Jimmy Ba.
\newblock Adam: A method for stochastic optimization.
\newblock In \emph{Proc. of ICLR}, 2015.

\bibitem[Kingma \& Welling(2014)Kingma and Welling]{kingma2014vae}
Diederik~P. Kingma and Max Welling.
\newblock Auto-encoding variational bayes.
\newblock In \emph{Proc. of ICLR}, 2014.

\bibitem[Kitaev et~al.(2020)Kitaev, Kaiser, and Levskaya]{kitaev2020reformer}
Nikita Kitaev, Lukasz Kaiser, and Anselm Levskaya.
\newblock Reformer: The efficient transformer.
\newblock In \emph{Proc. of ICLR}, 2020.

\bibitem[Lee et~al.(2019)Lee, Lee, Kim, Kosiorek, Choi, and Teh]{lee2019set}
Juho Lee, Yoonho Lee, Jungtaek Kim, Adam Kosiorek, Seungjin Choi, and Yee~Whye
  Teh.
\newblock Set transformer: A framework for attention-based
  permutation-invariant neural networks.
\newblock In \emph{Proc. of ICML}, 2019.

\bibitem[Li et~al.(2019)Li, Jin, Xuan, Zhou, Chen, Wang, and
  Yan]{li2019enhancing}
Shiyang Li, Xiaoyong Jin, Yao Xuan, Xiyou Zhou, Wenhu Chen, Yu-Xiang Wang, and
  Xifeng Yan.
\newblock Enhancing the locality and breaking the memory bottleneck of
  transformer on time series forecasting.
\newblock In \emph{Proc. of NeurIPS}, 2019.

\bibitem[Liu et~al.(2018)Liu, Saleh, Pot, Goodrich, Sepassi, Kaiser, and
  Shazeer]{liu2018generating}
Peter~J. Liu, Mohammad Saleh, Etienne Pot, Ben Goodrich, Ryan Sepassi, Lukasz
  Kaiser, and Noam Shazeer.
\newblock Generating wikipedia by summarizing long sequences.
\newblock In \emph{Proc. of ICLR}, 2018.

\bibitem[Maas et~al.(2011)Maas, Daly, Pham, Huang, Ng, and
  Potts]{maas2011learning}
Andrew~L. Maas, Raymond~E. Daly, Peter~T. Pham, Dan Huang, Andrew~Y. Ng, and
  Christopher Potts.
\newblock Learning word vectors for sentiment analysis.
\newblock In \emph{Proc. of ACL}, 2011.

\bibitem[Merity et~al.(2017)Merity, Xiong, Bradbury, and
  Socher]{merity2016pointer}
Stephen Merity, Caiming Xiong, James Bradbury, and Richard Socher.
\newblock Pointer sentinel mixture models.
\newblock In \emph{Proc. of ICLR}, 2017.

\bibitem[Merity et~al.(2018)Merity, Keskar, and Socher]{merity2018reg}
Stephen Merity, Nitish~Shirish Keskar, and Richard Socher.
\newblock {Regularizing and Optimizing LSTM Language Models}.
\newblock In \emph{Proc. of ICLR}, 2018.

\bibitem[Miconi et~al.(2018)Miconi, Stanley, and
  Clune]{thomas2018differentiable}
Thomas Miconi, Kenneth Stanley, and Jeff Clune.
\newblock Differentiable plasticity: training plastic neural networks with
  backpropagation.
\newblock In \emph{Proc. of ICML}, 2018.

\bibitem[Miculicich et~al.(2018)Miculicich, Ram, Pappas, and
  Henderson]{miculicich2018document}
Lesly Miculicich, Dhananjay Ram, Nikolaos Pappas, and James Henderson.
\newblock Document-level neural machine translation with hierarchical attention
  networks.
\newblock In \emph{Proc. of EMNLP}, 2018.

\bibitem[Mohamed et~al.(2019)Mohamed, Okhonko, and
  Zettlemoyer]{mohamed2019transformers}
Abdelrahman Mohamed, Dmytro Okhonko, and Luke Zettlemoyer.
\newblock Transformers with convolutional context for {ASR}.
\newblock \emph{arXiv: 1904.11660}, 2019.

\bibitem[Nangia \& Bowman(2018)Nangia and Bowman]{nangia2018listops}
Nikita Nangia and Samuel Bowman.
\newblock {L}ist{O}ps: A diagnostic dataset for latent tree learning.
\newblock In \emph{Proc. of NAACL Student Research Workshop}, 2018.

\bibitem[Oliva et~al.(2015)Oliva, Neiswanger, Poczos, Xing, Trac, Ho, and
  Schneider]{junier2015fast}
Junier Oliva, William Neiswanger, Barnabas Poczos, Eric Xing, Hy~Trac, Shirley
  Ho, and Jeff Schneider.
\newblock Fast function to function regression.
\newblock In \emph{Proc. of AISTATS}, 2015.

\bibitem[Ott et~al.(2018)Ott, Edunov, Grangier, and Auli]{ott2018scaling}
Myle Ott, Sergey Edunov, David Grangier, and Michael Auli.
\newblock Scaling neural machine translation.
\newblock In \emph{Proc. of WMT}, 2018.

\bibitem[Parisotto et~al.(2020)Parisotto, Song, Rae, Pascanu, Gulcehre,
  Jayakumar, Jaderberg, Kaufman, Clark, Noury, Botvinick, Heess, and
  Hadsell]{parisotto2019stabilizing}
Emilio Parisotto, H.~Francis Song, Jack~W. Rae, Razvan Pascanu, Caglar
  Gulcehre, Siddhant~M. Jayakumar, Max Jaderberg, Raphael~Lopez Kaufman, Aidan
  Clark, Seb Noury, Matthew~M. Botvinick, Nicolas Heess, and Raia Hadsell.
\newblock Stabilizing transformers for reinforcement learning.
\newblock In \emph{Proc. of ICML}, 2020.

\bibitem[Parmar et~al.(2018)Parmar, Vaswani, Uszkoreit, Kaiser, Shazeer, Ku,
  and Tran]{parmar2018image}
Niki Parmar, Ashish Vaswani, Jakob Uszkoreit, Lukasz Kaiser, Noam Shazeer,
  Alexander Ku, and Dustin Tran.
\newblock Image transformer.
\newblock In \emph{Proc. of ICML}, 2018.

\bibitem[Peng et~al.(2018)Peng, Schwartz, Thomson, and Smith]{peng2018rational}
Hao Peng, Roy Schwartz, Sam Thomson, and Noah~A. Smith.
\newblock Rational recurrences.
\newblock In \emph{Proc. of EMNLP}, 2018.

\bibitem[Peng et~al.(2020)Peng, Schwartz, Li, and Smith]{peng2020mixture}
Hao Peng, Roy Schwartz, Dianqi Li, and Noah~A. Smith.
\newblock A mixture of $h - 1$ heads is better than $h$ heads.
\newblock In \emph{Proc. of ACL}, 2020.

\bibitem[Post(2018)]{matt2014call}
Matt Post.
\newblock A call for clarity in reporting {BLEU} scores.
\newblock In \emph{Proc. of WMT}, 2018.

\bibitem[Qiu et~al.(2020)Qiu, Ma, Levy, Yih, Wang, and Tang]{qiu2019blockwise}
Jiezhong Qiu, Hao Ma, Omer Levy, Wen-tau Yih, Sinong Wang, and Jie Tang.
\newblock Blockwise self-attention for long document understanding.
\newblock In \emph{Findings of EMNLP}, 2020.

\bibitem[Radev et~al.(2009)Radev, Muthukrishnan, and Qazvinian]{radev2009aan}
Dragomir~R. Radev, Pradeep Muthukrishnan, and Vahed Qazvinian.
\newblock The {ACL} {A}nthology network.
\newblock In \emph{Proc. of the Workshop on Text and Citation Analysis for
  Scholarly Digital Libraries}, 2009.

\bibitem[Radford et~al.(2018)Radford, Wu, Child, Luan, Amodei, and
  Sutskever]{radford2018language}
Alec Radford, Jeffrey Wu, Rewon Child, David Luan, Dario Amodei, and Ilya
  Sutskever.
\newblock Language models are unsupervised multitask learners, 2018.

\bibitem[Rae et~al.(2020)Rae, Potapenko, Jayakumar, Hillier, and
  Lillicrap]{Rae2020Compressive}
Jack~W. Rae, Anna Potapenko, Siddhant~M. Jayakumar, Chloe Hillier, and
  Timothy~P. Lillicrap.
\newblock Compressive transformers for long-range sequence modelling.
\newblock In \emph{Proc. of ICLR}, 2020.

\bibitem[Rahimi \& Recht(2007)Rahimi and Recht]{rahimi2009rff}
Ali Rahimi and Benjamin Recht.
\newblock Random features for large-scale kernel machines.
\newblock In \emph{Proc. of NeurIPS}, 2007.

\bibitem[Rawat et~al.(2019)Rawat, Chen, Yu, Suresh, and
  Kumar]{rawat2019sampled}
Ankit~Singh Rawat, Jiecao Chen, Felix Xinnan~X Yu, Ananda~Theertha Suresh, and
  Sanjiv Kumar.
\newblock Sampled softmax with random {Fourier} features.
\newblock In \emph{Proc. of NeurIPS}, 2019.

\bibitem[Roy et~al.(2020)Roy, Saffar, Grangier, and Vaswani]{roy2020efficient}
Aurko Roy, Mohammad~Taghi Saffar, David Grangier, and Ashish Vaswani.
\newblock Efficient content-based sparse attention with routing transformers.
\newblock \emph{arXiv: 2003.05997}, 2020.

\bibitem[Sanh et~al.(2020)Sanh, Debut, Chaumond, and
  Wolf]{Sanh2019DistilBERTAD}
Victor Sanh, Lysandre Debut, Julien Chaumond, and Thomas Wolf.
\newblock {DistilBERT}, a distilled version of {BERT}: smaller, faster, cheaper
  and lighter.
\newblock \emph{arXiv: 1910.01108}, 2020.

\bibitem[{Schmidhuber}(1992)]{schmidhuber1992learning}
J.~{Schmidhuber}.
\newblock Learning to control fast-weight memories: An alternative to dynamic
  recurrent networks.
\newblock \emph{Neural Computation}, 4\penalty0 (1):\penalty0 131--139, 1992.

\bibitem[Schmidhuber(1993)]{schmidhuber1993reducing}
J.~Schmidhuber.
\newblock Reducing the ratio between learning complexity and number of time
  varying variables in fully recurrent nets.
\newblock In \emph{Proc. of ICANN}, 1993.

\bibitem[Sennrich et~al.(2016)Sennrich, Haddow, and Birch]{sennrich2016bpe}
Rico Sennrich, Barry Haddow, and Alexandra Birch.
\newblock Neural machine translation of rare words with subword units.
\newblock In \emph{Proc. of ACL}, 2016.

\bibitem[Shen et~al.(2020)Shen, Dong, Ye, Ma, Yao, Gholami, Mahoney, and
  Keutzer]{shen2020}
Sheng Shen, Zhen Dong, Jiayu Ye, Linjian Ma, Zhewei Yao, Amir Gholami,
  Michael~W. Mahoney, and Kurt Keutzer.
\newblock {Q-BERT}: {Hessian} based ultra low precision quantization of {BERT}.
\newblock In \emph{Proc. of AAAI}, 2020.

\bibitem[Sukhbaatar et~al.(2019)Sukhbaatar, Grave, Bojanowski, and
  Joulin]{sukhbaatar2019adaptive}
Sainbayar Sukhbaatar, Edouard Grave, Piotr Bojanowski, and Armand Joulin.
\newblock Adaptive attention span in transformers.
\newblock In \emph{Proc. of ACL}, 2019.

\bibitem[Sun(2019)]{sun2019random}
Yitong Sun.
\newblock \emph{Random Features Methods in Supervised Learning}.
\newblock PhD thesis, The University of Michigan, 2019.

\bibitem[Tay et~al.(2020{\natexlab{a}})Tay, Bahri, Metzler, Juan, Zhao, and
  Zheng]{tay2020synthesizer}
Yi~Tay, Dara Bahri, Donald Metzler, Da-Cheng Juan, Zhe Zhao, and Che Zheng.
\newblock Synthesizer: Rethinking self-attention in transformer models.
\newblock \emph{arXiv: 2005.00743}, 2020{\natexlab{a}}.

\bibitem[Tay et~al.(2020{\natexlab{b}})Tay, Bahri, Yang, Metzler, and
  Juan]{tay20sparse}
Yi~Tay, Dara Bahri, Liu Yang, Don Metzler, and Da-Cheng Juan.
\newblock Sparse sinkhorn attention.
\newblock In \emph{Proc. of ICML}, 2020{\natexlab{b}}.

\bibitem[Tay et~al.(2020{\natexlab{c}})Tay, Dehghani, Bahri, and
  Metzler]{tay2020}
Yi~Tay, Mostafa Dehghani, Dara Bahri, and Donald Metzler.
\newblock Efficient transformers: A survey.
\newblock \emph{arXiv: 2009.06732}, 2020{\natexlab{c}}.

\bibitem[Tay et~al.(2021)Tay, Dehghani, Abnar, Shen, Bahri, Pham, Rao, Yang,
  Ruder, and Metzler]{tay2020long}
Yi~Tay, Mostafa Dehghani, Samira Abnar, Yikang Shen, Dara Bahri, Philip Pham,
  Jinfeng Rao, Liu Yang, Sebastian Ruder, and Donald Metzler.
\newblock Long range arena: A benchmark for efficient transformers.
\newblock In \emph{Proc. of ICLR}, 2021.

\bibitem[Tsai et~al.(2019)Tsai, Bai, Yamada, Morency, and
  Salakhutdinov]{tsai2019transformer}
Yao-Hung~Hubert Tsai, Shaojie Bai, Makoto Yamada, Louis-Philippe Morency, and
  Ruslan Salakhutdinov.
\newblock Transformer dissection: An unified understanding for transformer{'}s
  attention via the lens of kernel.
\newblock In \emph{Proc. of EMNLP}, 2019.

\bibitem[Vaswani et~al.(2017)Vaswani, Shazeer, Parmar, Uszkoreit, Jones, Gomez,
  Kaiser, and Polosukhin]{vaswani2017attention}
Ashish Vaswani, Noam Shazeer, Niki Parmar, Jakob Uszkoreit, Llion Jones,
  Aidan~N Gomez, \L~ukasz Kaiser, and Illia Polosukhin.
\newblock Attention is all you need.
\newblock In \emph{Proc. of NeurIPS}, 2017.

\bibitem[Wang et~al.(2020)Wang, Li, Khabsa, Fang, and Ma]{wang2020linformer}
Sinong Wang, Belinda~Z. Li, Madian Khabsa, Han Fang, and Hao Ma.
\newblock Linformer: Self-attention with linear complexity.
\newblock \emph{arXiv: 2006.04768}, 2020.

\bibitem[Williams \& Zipser(1989)Williams and Zipser]{williams1989recurrent}
Ronald~J. Williams and David Zipser.
\newblock A learning algorithm for continually running fully recurrent neural
  networks.
\newblock \emph{Neural Computation}, 1:\penalty0 270--280, 1989.

\bibitem[Wu et~al.(2019)Wu, Fan, Baevski, Dauphin, and Auli]{wu2018pay}
Felix Wu, Angela Fan, Alexei Baevski, Yann Dauphin, and Michael Auli.
\newblock Pay less attention with lightweight and dynamic convolutions.
\newblock In \emph{Proc. of ICLR}, 2019.

\bibitem[Wu et~al.(2020)Wu, Liu, Lin, Lin, and Han]{wu2020Lite}
Zhanghao Wu, Zhijian Liu, Ji~Lin, Yujun Lin, and Song Han.
\newblock Lite transformer with long-short range attention.
\newblock In \emph{Proc. of ICLR}, 2020.

\bibitem[You et~al.(2020)You, Sun, and Iyyer]{you2020hard}
Weiqiu You, Simeng Sun, and Mohit Iyyer.
\newblock Hard-coded {G}aussian attention for neural machine translation.
\newblock In \emph{Proc. of ACL}, 2020.

\bibitem[Yu et~al.(2016)Yu, Suresh, Choromanski, Holtmann-Rice, and
  Kumar]{yu2016orf}
Felix Xinnan~X Yu, Ananda~Theertha Suresh, Krzysztof~M Choromanski, Daniel~N
  Holtmann-Rice, and Sanjiv Kumar.
\newblock Orthogonal random features.
\newblock In \emph{Proc. of NeurIPS}, 2016.

\bibitem[Zaheer et~al.(2020)Zaheer, Guruganesh, Dubey, Ainslie, Alberti,
  Ontanon, Pham, Ravula, Wang, Yang, and Ahmed]{zaheer2020big}
Manzil Zaheer, Guru Guruganesh, Avinava Dubey, Joshua Ainslie, Chris Alberti,
  Santiago Ontanon, Philip Pham, Anirudh Ravula, Qifan Wang, Li~Yang, and Amr
  Ahmed.
\newblock Big bird: Transformers for longer sequences.
\newblock \emph{arXiv: 2007.14062}, 2020.

\end{thebibliography}
\bibliographystyle{iclr2021_conference}

\clearpage
\appendix
\begin{appendices}

\section{Random Feature Attention in More Detail}\label{sec:rfa_detail}
\subsection{Detailed Computation Procedure}\label{sec:algo} 
Algorithms~\ref{algo:rfa_causal} and~\ref{algo:rfa_cross}
describe causal and cross random feature
attention's computation procedures.

\begin{figure*}[ht]
\centering
\begin{minipage}{.65\linewidth}
\begin{algorithm}[H]
	\centering
	\caption{Causal random feature attention.}
	\label{algo:rfa_causal}
	\begin{algorithmic}[1]
		\Procedure{RFA-Causal}{
		    $\{\vq_i\}_{i=1}^{N}$, $\{\vk_i\}_{i=1}^{N}$, $\{\vv_i\}_{i=1}^{N}$}
		\State\Comment $\mS$ is a $D\times d$ matrix
		\State\Comment $\vz$ is a $D$-dimensional vector 
		\State $\mS, \, \vz\leftarrow \vzero,\,\vzero$ 
		\For{$i=1$ \textbf{to} $N$}
		\State $\widetilde{\vq}_i, \, \widetilde{\vk}_i \leftarrow \boldsymbol{\phi}(\vq_i), \, \boldsymbol{\phi}(\vk_i)$
		\quad\Comment Random feature maps
		\State $\mS \leftarrow \mS + \widetilde{\vk}_i\otimes\vv_i$
		\State $\vz \leftarrow \vz + \widetilde{\vk}_i$
		\State $\vh_i^\top\leftarrow {\widetilde{\vq}_i^\top\mS} / {(\widetilde{\vq}_i\cdot\vz)}$
		\EndFor
		\State\Return $\{\vh_i\}_{i=1}^{N}$
		\EndProcedure
	\end{algorithmic}
\end{algorithm}
\end{minipage}
\end{figure*}
\begin{figure*}[ht]
\centering
\begin{minipage}{.6\linewidth}
\begin{algorithm}[H]
	\centering
	\caption{Cross random feature attention.}
	\label{algo:rfa_cross}
	\begin{algorithmic}[1]
		\Procedure{RFA-Cross}{
		    $\{\vq_i\}_{i=1}^{N}$, 
		    $\{\vk_i\}_{i=1}^{M}$,
		    $\{\vv_i\}_{i=1}^{M}$}
		\State\Comment $\mS$ is a $D\times d$ matrix
		\State\Comment $\vz$ is a $D$-dimensional vector 
		\State $\mS, \, \vz\leftarrow \vzero,\,\vzero$ 
		\For{$i=1$ \textbf{to} $M$}
		\State $\widetilde{\vk}_i \leftarrow \boldsymbol{\phi}(\vk_i)$
		\quad\Comment Random feature map
		\State $\mS \leftarrow \mS + \widetilde{\vk}_i\otimes\vv_i^\top$
		\State $\vz \leftarrow \vz + \widetilde{\vk}_i$
		\EndFor
		
		\For{$i=1$ \textbf{to} $N$}
		\State $\widetilde{\vq}_i \leftarrow \boldsymbol{\phi}(\vq_i)$
		\quad\Comment Random feature map
		\State $\vh_i^\top\leftarrow {\widetilde{\vq}_i^\top\mS} / {(\widetilde{\vq}_i\cdot\vz)}$
		\label{algo:line:cum_s}
		\EndFor
		\State\Return $\{\vh_i\}_{i=1}^{N}$
		\EndProcedure
	\end{algorithmic}
\end{algorithm}
\end{minipage}
\end{figure*}

\rev{\subsection{Variance of Random Fourier Features}\label{sec:variance}}
\rev{The following result is due to~\citet{yu2016orf}.
Using the same notation as in~\S\ref{sec:random_feature}:
\begin{align}
\operatorname{Var}(\boldsymbol{\phi}\left(\vx\right)\cdot \boldsymbol{\phi}\left(\vy\right))
=\frac{1}{2D}\left(1 - e^{-z^2}\right)^2,
\end{align}
where $z=\norm{\vx-\vy} / \sigma$.
}

\rev{
\subsection{Derivation of Causal Rfa}\label{sec:causal_rfa_derivation}}
\rev{This section presents a detailed derivation of causal \model as in \S\ref{sec:rfa}.
Following Eq.~\ref{eq:approx_softmax} but changing the attended keys and values to the prefix:
}
\rev{
\begin{align}\label{eq:causal_rfa_derivation}
    \model(\vq_t,\{\vk_i\}_{i\leq t},\{\vv_i\}_{i\leq t}) 
    &= \frac{\boldsymbol{\phi}\left(\vq_t\right)^\top\sum_{i\leq t} \boldsymbol{\phi}\left(\vk_i\right)\otimes\vv_i}
    {\boldsymbol{\phi}\left(\vq_t\right)\cdot\sum_{j\leq t} \boldsymbol{\phi}\left(\vk_j\right)}
\end{align}
Let $\mS_t \triangleq \sum_{i\leq t} \boldsymbol{\phi}(\vk_i)\otimes\vv_i$,
and $\vz_t \triangleq  \sum_{i\leq t} \boldsymbol{\phi}(\vk_i)$;
both can be calculated recurrently.
Assuming $\mS_0 = \mathbf{0}$ and $\vz_0 = \mathbf{0}$:
\begin{align}
    \mS_t = \mS_{t-1} +  \boldsymbol{\phi}\left(\vk_t\right)\otimes\vv_t, \quad
    \vz_t = \vz_{t-1} + \boldsymbol{\phi}\left(\vk_t\right), \quad t\geq 1.
\end{align}
This completes the derivation of causal \model as in \S\ref{sec:rfa}.
}

\subsection{Rfa without Norm-1 Constraints}\label{sec:no_norm_constraint}
\S\ref{sec:rfa} assumes that the queries and keys are unit vectors.
This norm-1 constraint is \emph{not} a must.
Here we present a \model \emph{without} imposing this constraint.
Let $C(\vx) = \exp(\norm{\vx}^2/2\sigma^2)$.
From Eq.~\ref{eq:approx_dot_exp} we have
$\operatorname{attn}\left(\vq_t,\{\vk_i\},\{\vv_i\}\right)=$
\begin{align}\label{eq:approx_softmax_norm}
\begin{split}
    \sum_i
    \frac{\exp\left(\vq_t\cdot\vk_i /\sigma^2\right)}
    {\sum_j\exp\left(\vq_t\cdot\vk_j /\sigma^2\right)}\vv_i^\top
    &\approx 
    \sum_i\frac{C(\vq_t)\,C(\vk_i)\,\boldsymbol{\phi}\left(\vq_t\right)^\top \boldsymbol{\phi}\left(\vk_i\right)\vv_i^\top}
    {\sum_j C(\vq_t)\,C(\vk_j)\,\boldsymbol{\phi}\left(\vq_t\right) \cdot \boldsymbol{\phi}\left(\vk_j\right)}\\
    &= \frac{\boldsymbol{\phi}\left(\vq_t\right)^\top\sum_i C(\vk_i)\,\boldsymbol{\phi}\left(\vk_i\right)\otimes\vv_i}
    {\boldsymbol{\phi}\left(\vq_t\right)\cdot\sum_j 
    C(\vk_j)\,\boldsymbol{\phi}\left(\vk_j\right)}.
\end{split}
\end{align} 
The specific attention computation is similar to those in \S\ref{sec:rfa}.
In sum, lifting the norm-1 constraint
brings an additional scalar term $C(\boldsymbol{\cdot})$.

\rev{
\subsection{Relating Rfa-Gate to Softmax Attention}\label{sec:gated_softmax}
Drawing inspiration from gated RNNs,
\S\ref{sec:gate} introduces a gated variant of \model.
Now we study its ``softmax counterpart.''
\begin{align}
\begin{split}
\widetilde{\mathbf{k}}_i &= \mathbf{k}_i \  (1-g_i) \prod_{j=i+1}^{t}g_j,
\quad \widetilde{\mathbf{v}}_i = \mathbf{v}_i  \ (1-g_i) \prod_{j=i+1}^{t}g_j, \quad i=1,\dots,t \\
\mathbf{h}_t&=\operatorname{attn}(\mathbf{q}_t, \{\widetilde{\mathbf{k}}_i\}_{i\leq t}, \{\widetilde{\mathbf{v}}_i\}_{i\leq t}).
\end{split}
\end{align}
$\mathbf{h}_t$ is the output at timestep $t$ and is used for onward computation.
}

\rev{
At each step, all prefix keys and values are decayed by a gate value before calculating the attention.
This implies that the attention computation for $\mathbf{q}_{t+1}$ \emph{cannot} start until that of $\mathbf{q}_t$ is finished. 
Combined with the linear complexity of softmax normalization, this amounts to quadratic time in sequence length, 
even for language modeling training. 
}

\rev{
The above model is less intuitive and more expensive in practice, without the \textsc{Rfa} perspective. This shows that \model brings some benefits in developing new attention models.
}

\subsection{Detailed Complexity Analysis}\label{sec:app:complexity}
Table~\ref{tab:complexity} considers 
a sequence-to-sequence model,
and breaks down the comparisons
to training (with teacher forcing;~\citealp{williams1989recurrent})
and autoregressive decoding.
\rev{Here we assume enough threads to fully parallelize
softmax attention across timesteps when the inputs are revealed to the model in full.}
\model has a lower space complexity, 
since it never explicitly populates the attention matrices.
As for time, 
\model trains in linear time,
and so does the softmax attention:
in teacher-forcing training 
a standard transformer decoder parallelizes the attention computation 
across time steps.
The trend of the time comparison differs during decoding:
when only one output token is produced at a time,
\model decodes linearly in the output length,
while softmax attention decodes  quadratically.

\begin{table*}[th]
\centering

    \begin{tabular}{@{} ll c@{\hspace{5pt}}c@{\hspace{5pt}}c m{0.001em} c@{\hspace{5pt}}c@{\hspace{5pt}}c@{}}
    \toprule[.1em]
    &&\multicolumn{3}{c}{\textbf{Time Complexity}}
    &&\multicolumn{3}{c}{\textbf{Space Complexity}}\\
    \cmidrule(lr){3-5}
    \cmidrule(lr){7-9}
    \textbf{Setting} & \textbf{Model}  
    & \textbf{Encoder} & \textbf{Cross} & \textbf{Causal}
    && \textbf{Encoder} & \textbf{Cross} & \textbf{Causal}\\
    \midrule[.1em]
    \multirow{2}{*}{\shortstack[l]{Training w/ \\teacher forcing}}
    & softmax
    & $\gO(M)$ & $\gO(M)$ & $\gO(N)$ 
    && $\gO(M^2)$ & $\gO(MN)$ & $\gO(N^2)$\\
    & \model
    & $\gO(M)$ & $\gO(M)$ & $\gO(N)$
    && ${\color{myblue}\gO(M)}$ & ${\color{myblue}\gO(M+N)}$ & ${\color{myblue}\gO(N)}$\\
    
    \midrule
    \multirow{2}{*}{Decoding}
    & softmax & $\gO(M)$ & $\gO(MN)$ &  $\gO(N^2)$
    && $\gO(M^2)$ & $\gO(MN)$ & $\gO(N^2)$\\
    & \model & $\gO(M)$ & ${\color{myblue}\gO(M+N)}$ & ${\color{myblue}\gO(N)}$
    && ${\color{myblue}\gO(M)}$ & ${\color{myblue}\gO(M+N)}$ & ${\color{myblue}\gO(N)}$\\
    
    \bottomrule[.1em]
    \end{tabular}

\caption{Time and space complexity comparisons between \model and its softmax counterpart in a sequence-to-sequence attentive model,
\rev{assuming an infinite amount of available threads}.
$M$ and $N$ denote the lengths of the source and target sequences respectively.
Teacher forcing training~\citep{williams1989recurrent} and autoregressive decoding are assumed. 
{\color{myblue}Blue color} 
indicates the cases where \model asymptotically outperforms softmax attention. } 
\label{tab:complexity}
\end{table*}

\begin{table*}[th]
\centering
	\centering
	\begin{tabular}{@{}l  cccc@{}} 
		\toprule
		
		\textbf{Data} & \textbf{Train} & \textbf{Dev.} & \textbf{Test} & \textbf{Vocab.}\\ 
		
		\midrule
		WikiText-103 & 103M & 218K & 246K & 268K\\
		\midrule
		
		WMT14 EN-DE & 4.5M & 3K & 3K & \phantom{/K}32K\\
		WMT14 EN-FR & 4.5M & 3K & 3K & \phantom{/K}32K\\
		IWSLT14 DE-EN & 160K& 7K& 7K& 9K/7K\\
		
		\bottomrule
	\end{tabular}
	\caption{Some statistics for the datasets.
	    WikiText-103 split sizes are in number of tokens,
	    while others are in number of instances.
	}
	\label{tab:data}
\end{table*}

\section{Experimental Details}\label{sec:experimental_details}
Table~\ref{tab:data} summarizes
some statistics of the datasets used in our experiments.
Our implementation is based on JAX.\footnote{
\url{https://github.com/google/jax}.
}

\begin{table*}[th]
\centering
\begin{tabular}{@{} l cccc@{}}
\toprule[.1em]

\textbf{\# Random Matrices} & 1 & 50 & 100 & 200 \\
\midrule[.1em]
\textbf{BLEU} & 24.0 & 25.7 & 25.8 & 25.8\\
\bottomrule[.1em]
\end{tabular}
\caption{WMT14 EN-DE development set performance varying the 
number of random matrices to sample from during training.
No beam search or checkpoint averaging is used.} \label{tab:number_random_matrices}
\end{table*}

\rev{During training, we sample a different random projection matrix for each attention head.
Preliminary experiments suggest this performs better than using the same random projection throughout training~(Table~\ref{tab:number_random_matrices}).
Our conjecture is that this helps keep the attention heads from ``over committing'' to any particular random projection~\citep{peng2020mixture}.
To avoid the overhead of sampling from Gaussian during training,
we do this in an offline manner.
I.e., before training we construct a pool of random matrices (typically 200),
at each training step we draw from the pool.
At test time each attention head uses the same random projection, since no accuracy benefit is observed by using different ones for different test instances.
}
\subsection{Language Modeling}\label{sec:lm_details}
We compare the models using two model size settings, summarized in
Table~\ref{tb:lm_setting}.
We use the fixed sinusoidal position embeddings by~\citet{vaswani2017attention}.
All models are trained for up to 150K gradient steps
using the Adam optimizer~\citep{kingma2014adam}.
No $\ell_2$-regularization is used.
We apply early stopping based on development set perplexity.
All models are trained using 16 TPU v3 accelerators,
and tested using a single TPU v2 accelerator.

\begin{table*}[th]
\centering
\begin{tabular}{@{} l cc@{}}
\toprule[.1em]

\textbf{Hyperprams.} & \textbf{Small} & \textbf{Big}\\

\midrule[.1em]
\# Layers & 6 & 16\\
\# Heads & 8 & 16\\
Embedding Size  & 512 & 1024\\
Head Size & 64 & 64\\
FFN Size & 2048 & 4096\\
Batch Size & 64 & 64\\
Learning Rate & \multicolumn{2}{c}{$[1\times10^{-4}, 2.5\times10^{-4}, 5\times10^{-4}]$}\\
Warmup Steps & 6000 & 6000\\
Gradient Clipping Norm & 0.25 & 0.25\\
Dropout &[0.05, 0.1]  & [0.2, 0.25, 0.3]\\
Random Feature Map Size & 64 & 64\\

\bottomrule[.1em]
\end{tabular}
\caption{\label{tb:lm_setting}
Hyperparameters used in the language modeling experiments.} 
\end{table*}

\subsection{Machine Translation}\label{sec:mt_details}
\textbf{WMT14.}
We use the fixed sinusoidal position embeddings by~\citet{vaswani2017attention}.
For both EN-DE and EN-FR experiments,
we train the models using the Adam (with $\beta_1=0.1$, $\beta_2=0.98$, and $\epsilon=10^{-9}$) optimizer
for up to 350K gradient steps.
We use a batch size of 1,024 instances for EN-DE,
while 4,096 for the much larger EN-FR dataset.
The learning rate follows that by \citet{vaswani2017attention}.
Early stopping is applied based on development set BLEU.
No $\ell_2$ regularization or gradient clipping is used.
All models are trained using 16 TPU v3 accelerators,
and tested using a single TPU v2 accelerator.
Following standard practice, we average 10 most recent checkpoints at test time.
We evaluate the models using SacreBLEU~\citep{matt2014call}.\footnote{\url{https://github.com/mjpost/sacrebleu}}
A beam search with beam size 4 and length penalty 0.6 is used.
Other hyperparameters are summarized in Table~\ref{tb:mt_hyperparams}.

\begin{table*}[th]
\centering
\begin{tabular}{@{} l cc@{}}
\toprule[.1em]

\textbf{Hyperprams.} & \textbf{WMT14} & \textbf{IWSLT14}\\

\midrule[.1em]
\# Layers & 6 & 6\\
\# Heads & 8 & 8\\
Embedding Size  & 512 & 512\\
Head Size & 64 & 64\\
FFN Size & 2048 & 2048\\
Warmup Steps & 6000 & 4000\\
Dropout & 0.1 & 0.3\\
Cross Attention Feature Map & 128 & 128\\
Causal Attention Feature Map & 64 & 64\\

\bottomrule[.1em]
\end{tabular}
\caption{\label{tb:mt_hyperparams}
Hyperparameters used in the machine translation experiments.} 
\end{table*}

\section{More Analysis Results}
\subsection{More Results on Decoding Speed and Memory Overhead}\label{sec:more_length}
\begin{figure*}
\centering
\begin{subfigure}[tb]{0.49\textwidth}
\includegraphics[trim={.5cm 0cm 2.5cm 0cm},clip,width=\textwidth]{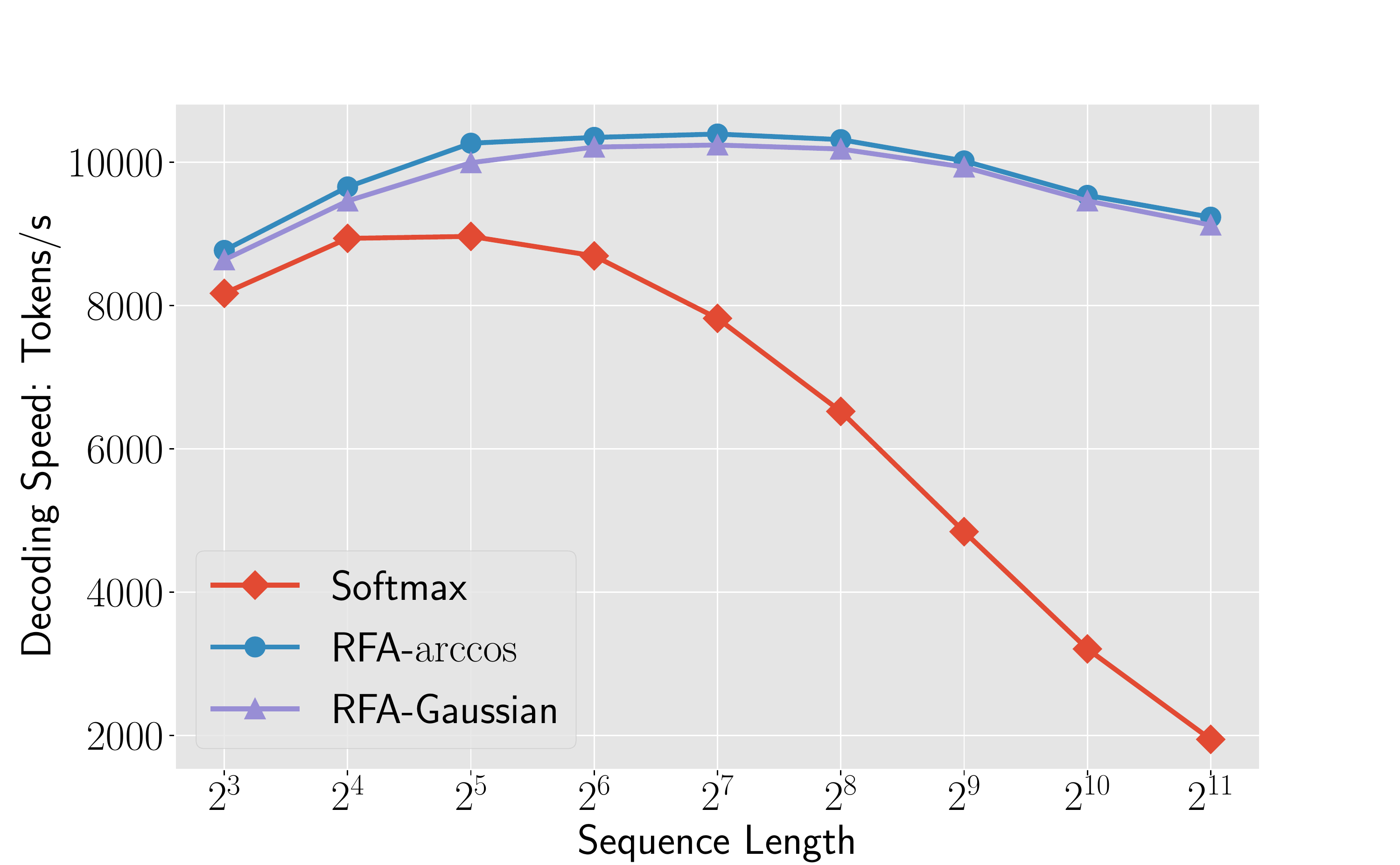}
\caption{Speed vs. lengths.}
\end{subfigure}
\begin{subfigure}[tb]{0.49\textwidth}
\includegraphics[trim={.5cm 0cm 2.5cm 0cm},clip,width=\textwidth]{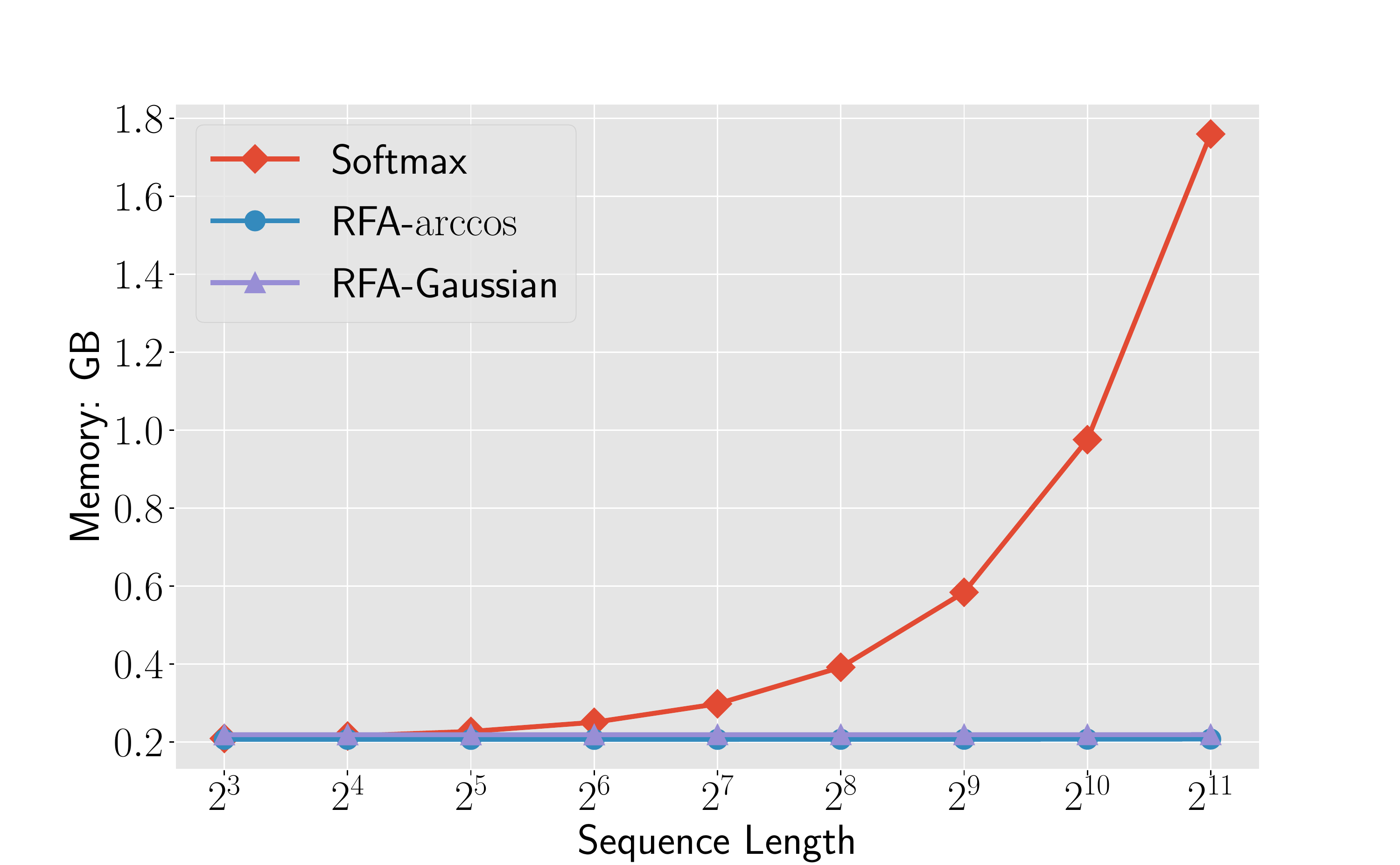}
\caption{\label{fig:length_memory_unconditional}Memory vs. lengths.}
\end{subfigure}
\caption{Unconditional decoding speed (left) and memory overhead (right) varying the output lengths. All models are tested on a single TPU v2 accelerator, with greedy decoding and batch size 16.}\label{fig:length_unconditional}
\end{figure*}
Figure~\ref{fig:length_unconditional}
compares the \model's \emph{unconditional} decoding speed and memory 
against the softmax attention.
The setting is the same as that in \S\ref{sec:analysis}
except that here the models do not have an encoder.
This experiment aims to simulate the applications
such as sampling from a language model.

\rev{\subsection{Effect of Random Feature Size}\label{sec:random_feature_size}}
\rev{This section studies how the size of $\boldsymbol{\phi}(\boldsymbol{\cdot})$ affects the performance.}
\rev{Table~\ref{tab:random_feature_size} summarize \model-Gaussian's performance on WMT14 EN-DE development set.
The model and training are the same as that used in \S\ref{sec:mt} except random feature size.
Recall from \S\ref{sec:random_feature} that the size of $\boldsymbol{\phi}(\boldsymbol{\cdot})$ is $2D$ for \model-Gaussian.
When the size of $\boldsymbol{\phi}(\boldsymbol{\cdot})$ is too small (32 or 64 for cross attention, 32 for causal attention),
training does not converge. 
We observe accuracy improvements by using random features sufficiently large (256 for cross attention and 128 for causal attention); 
going beyond that, the benefit is marginal.}

\begin{table*}[th]
\begin{subtable}[tbh]{.48\textwidth}
\centering
\begin{tabular}{@{} l ccccc @{}}
\toprule[.1em]

$\boldsymbol{\phi}$ \textbf{Size} & 32 & 64 & 128 & 256 & 512\\
\midrule[.1em]
\textbf{BLEU} & N/A & N/A & 24.9 & 25.8 & 26.0\\
\bottomrule[.1em]
\end{tabular}
\caption{Varying cross attention $\boldsymbol{\phi}$ sizes
while fixing that of causal attention to be 128.} 
\end{subtable}
\hfill
\begin{subtable}[tbh]{.48\textwidth}
\centering
\begin{tabular}{@{} l ccccc @{}}
\toprule[.1em]

$\boldsymbol{\phi}$ \textbf{Size} & 32 & 64 & 128 & 256 & 512\\
\midrule[.1em]
\textbf{BLEU} & N/A & 25.3 & 25.8 & 25.8 & 25.6\\
\bottomrule[.1em]
\end{tabular}
\caption{Varying causal attention $\boldsymbol{\phi}$ sizes
while fixing that of cross attention to be 256.} 
\end{subtable}
\caption{WMT14 EN-DE development set performance of \model-Gaussian 
(the size of $\boldsymbol{\phi}$ is $2D$; \S\ref{sec:random_feature})
varying the random feature sizes.
N/A indicates training does not converge.
No beam search or checkpoint averaging is used.
} \label{tab:random_feature_size}
\end{table*}

\subsection{Train and Evaluate with Different Attention Functions}
\label{sec:train_eval}
\model achieves comparable performance to its softmax counterpart.
Does this imply that it learns 
a good approximation to the softmax attention?
To answer this question, we consider:
\begin{compactitem}
\item[(i)] an \model-Gaussian model initialized 
from a pretrained softmax-transformer;
\item[(ii)] a softmax-transformer initialized
from a pretrained an \model-Gaussian model.
\end{compactitem}

If \model's good performance can be attributed
to learning a good approximation to softmax, 
both, \emph{without} finetunining,
should perform similarly to the pretrained models.
However, this is \emph{not} the case on IWSLT14 DE-EN.
Both pretrained models achieve more than 35.2 development set BLEU.
In contrast, (i) and (ii) respectively get 2.3 and 1.1 BLEU \emph{without} finetuning,
hardly beating a randomly-initialized untrained model.
This result aligns with the observation by~\citet{choromanski2020masked},
and suggests that it is \emph{not} the case 
that \model performs well because it learns
to imitate softmax attention's outputs.

\begin{figure}
\centering
\includegraphics[trim={.5cm 0cm 2.5cm 0cm},clip,width=.48\linewidth]{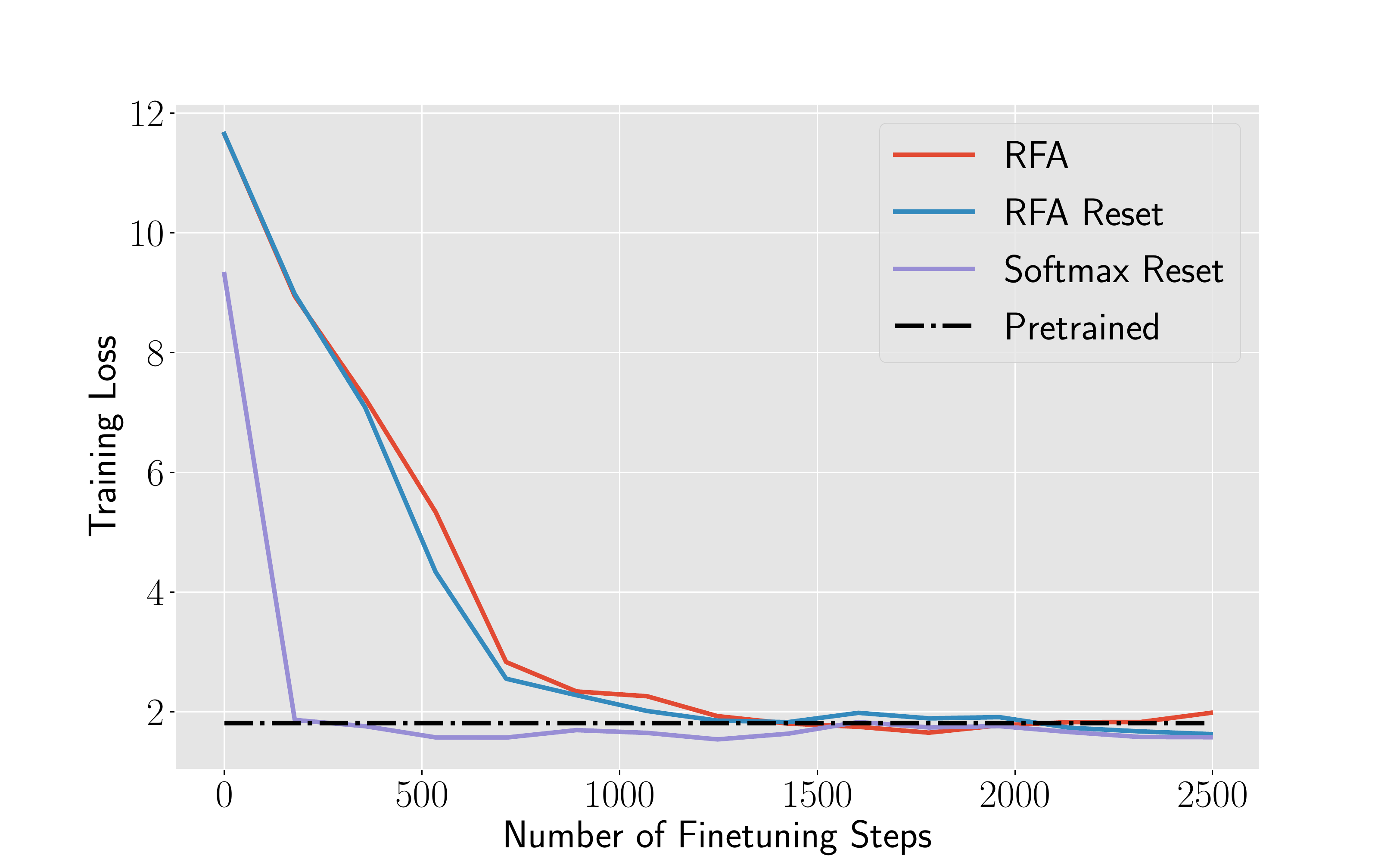}
\captionof{figure}{\label{fig:finetune}
Finetuning an \model-Gaussian model with its parameters initialized 
from a pretrained softmax-transformer.
``Reset'' indicates resetting the multihead attention parameters
to randomly-initialized ones.
The dashed line indicates the training loss of the pretrained model.
}
\end{figure}

\subsection{Knowledge Transfer from Softmax Attention to RFA}
\label{sec:finetune}
We first supplement the observation in Appendix~\ref{sec:train_eval}
by finetuning (i) on the same pretraining data.
Figure~\ref{fig:finetune} plots the learning curves.
It takes \model roughly 1,500 steps to reach similar training
loss to the pretrained model.
As a baseline, ``\model Reset''
resets the multihead attention parameters 
(i.e., those for query, key, value, and output projections)
to randomly initialized ones.
Its learning curve is similar to that of (i),
suggesting that the pretrained multihead attention parameters
are no more useful to \model than randomly initialized ones.
To further confirm this observation,
``softmax Reset'' resets the multihead attention parameters
\emph{without} changing the attention functions.
It converges to the pretraining loss in less than 200 steps.

\textbf{Takeaway.}
From the above results on IWSLT14,
pretrained knowledge in a softmax transformer
\emph{cannot} be directly transferred to an \model model.
However, from Figure~\ref{fig:finetune} and 
a much larger-scale experiment by \citet{choromanski2020masked},
we do observe that 
\model can recover the pretraining loss,
and the computation cost of finetuning 
is much less than training a model from scratch.
This suggests some potential applications.
For example, one might be able to initialize an \model language model 
from a softmax transformer pretrained on large-scale data 
(e.g., GPT-3;~\citealp{brown2020language}),
and finetune it at a low cost.
The outcome would be an \model model retaining most of the pretraining knowledge,
but is much faster and more memory-friendly to sample from.
We leave such exploration to future work.

\end{appendices}

\end{document}